%% file: main_arxiv.tex
\documentclass[10pt,twocolumn,letterpaper]{article}

\usepackage{iccv}
\usepackage{times}
\usepackage{epsfig}
\usepackage{graphicx}
\usepackage{amsmath}
\usepackage{amssymb}
\usepackage{float}
\usepackage{cuted}

\usepackage{color}
\usepackage{amssymb,amsmath}
\usepackage{multirow}
\usepackage{ctable}
\usepackage{array}
\usepackage{subcaption}

\usepackage{ulem}
\renewcommand{\arraystretch}{1.3}
\definecolor{Crimson}{rgb}{0.86, 0.08, 0.24}
\definecolor{DarkGreen}{rgb}{0.00, 0.40, 0.00}
\definecolor{RoyalBlue}{rgb}{0.15, 0.25, 0.54}
\definecolor{DarkCyan}{rgb}{0.0, 0.54, 0.54}
\ifx \submission \undefined

\else

\fi
\newcommand{\comment}[1]{}
\usepackage{makecell}

\usepackage{xspace}
\makeatletter
\DeclareRobustCommand\onedot{\futurelet\@let@token\@onedot}
\def\@onedot{\ifx\@let@token.\else.\null\fi\xspace}
\def\eg{\emph{e.g}\onedot} 
\def\ie{\emph{i.e}\onedot} 
 
 \def\vs{\emph{vs}\onedot}
 
\def\etal{\emph{et al}\onedot}
\makeatother
\usepackage{bm}

\usepackage{mathtools}
\DeclarePairedDelimiterX{\norm}[1]{\lVert}{\rVert}{#1}

\usepackage{bm}

\usepackage[pagebackref=true,breaklinks=true,letterpaper=true,colorlinks,bookmarks=false]{hyperref}

\iccvfinalcopy 

\def\iccvPaperID{259} 

\ificcvfinal\fi
\begin{document}

\newcommand{\modelname}{\text{Flat2Layout}\xspace}
\title{\modelname: Flat Representation for Estimating Layout of General Room Types}

\author{
Chi-Wei Hsiao \qquad Cheng Sun \qquad Min Sun \qquad Hwann-Tzong Chen \\
National Tsing Hua University \\
{\tt\small \{chiweihsiao, chengsun\}@gapp.nthu.edu.tw} \qquad {\tt\small sunmin@ee.nthu.edu.tw} \qquad {\tt\small htchen@cs.nthu.edu.tw}
}
\date{}

\twocolumn[{%
\maketitle
\renewcommand\twocolumn[1][]{#1}%
   \vspace{-5mm} 
    \centering
    \includegraphics[width=.85\linewidth]{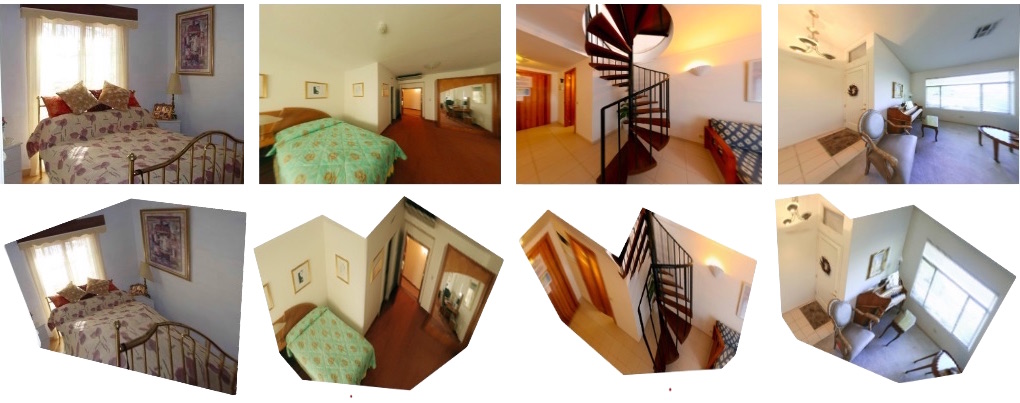}
    \captionof{figure}{Sample 3D room layouts reconstructed by our \modelname.\label{fig:teaser}}
   \vspace{5mm} 
}]

\begin{abstract}
This paper proposes a new approach, \modelname, for estimating general indoor room layout from a single-view RGB image whereas existing methods can only produce layout topologies captured from the box-shaped room. The proposed flat representation encodes the layout information into row vectors which are treated as the training target of the deep model. A dynamic programming based post-processing is employed to decode the estimated flat output from the deep model into the final room layout. \modelname achieves state-of-the-art performance on existing room layout benchmark. This paper also constructs a benchmark for validating the performance on general layout topologies, where \modelname achieves good performance on general room types. \modelname is applicable on more scenario for layout estimation and would have an impact on applications of Scene Modeling, Robotics, and Augmented Reality.
\end{abstract}

\input{sec-intro.tex}

\input{sec-relatedwork.tex}

\input{sec-approach.tex}

\input{sec-experiments.tex}

\input{sec-conclusion.tex}

{\small
\bibliographystyle{ieee}
\bibliography{main_arxiv}
}

\newpage
\begin{strip}
    \null
    \vskip .375in
    \begin{center}
      {\Large \bf Supplemental Material \par}
      \vspace*{24pt}
      {}
      \vskip .5em
      \vspace*{12pt}
    \end{center}
\end{strip}

\setcounter{section}{0}
\renewcommand\thesection{\Alph{section}}
\input{supp_lsun_types.tex}

\input{supp_derivation_3d.tex}

\input{supp_qual_hedau.tex}

\newpage
\onecolumn
\input{supp_qual_lsun.tex}

\newpage
\input{supp_qual_general.tex}

\end{document}

%% file: sec-intro.tex
\section{Introduction}
Estimating room layout is a fundamental indoor scene understanding problem with applications to a wide range of tasks such as scene reconstruction~\cite{lee2017joint}, indoor localization~\cite{boniardi2019robot} and augmented reality. Consider a single-view RGB image: the layout estimation task is to delineate the wall-ceiling, wall-floor, and wall-wall boundaries. Existing works only target special cases of room layouts that comprise at most five planes (\ie, ceiling, floor, left wall, front wall, and right wall).

Previous deep learning based methods \cite{dasgupta2016delay,mallya2015learning,ren2016coarse} typically predict 2D per-pixel edge maps or segmentation maps, (\ie ceiling, floor, left, front, and right), followed by the classic vanishing point/line sampling methods to produce room layouts.
However, none of these methods could directly apply to non-box-shaped room layout topology.
For instance, more segmentation labels have to be defined in the framework proposed by Dasgupta~\etal \cite{dasgupta2016delay} to generate a layout for a room which contains more than three walls.
In addition, these methods highly depend on the accuracy of the extraction of the three mutually orthogonal vanishing points, which sometimes fails due to misleading texture.

We propose \modelname, a layout estimation approach that can directly work for general topologies of room layouts without the need to extract horizontal vanishing points.
The overall pipeline includes vertical-axis image rectification, prediction of the proposed flat output representation with the proposed deep model, and an efficient post-processing procedure based on dynamic programming.

We summarize our contributions as follows:
\begin{itemize}
    \item We design a flat target output representation that could be efficiently decoded to layout (\ie corners or plane segmentation) with an intuitive and effective post-processing procedure based on dynamic programming.
    \item Our approach achieves state-of-the-art performance on the existing dataset with processing time as less as 500ms per frame (there is still room for speedup).
    \item Our method could apply to not only the typical 11 types defined in LSUN~\cite{yu15lsun} dataset but also more complex room layout topologies. We quantitatively and qualitatively demonstrate that our approach is capable of recovering room layout of general layout types from a single-view RGB image.
\end{itemize}

%% file: sec-relatedwork.tex
\section{Related Work}
Single-view room layout estimation has been an active task over the past decade.
Hedau \etal~\cite{hedau2009recovering} first defined the problem as using a cuboid-shaped box to approximate the 3D layout of indoor scenes. 
In their method, many box layout proposals were generated by sampling rays from the three orthogonal vanishing points. Then they ranked the candidates with a structured SVM trained on images annotated with surface labels \{left wall, front wall, right wall, ceiling, object\}.
Using a similar framework, many works explored different methods from the aspect of proposal generation~\cite{schwing2012efficient,schwing2013box,ramalingam2013manhattan} and inference procedure~\cite{lee2009geometric,hedau2010thinking,schwing2012efficient,ramalingam2013manhattan}.

Recently, per-pixel 2D target output representations were generally adopted in many deep-learning based approaches.
Mallya \etal~\cite{mallya2015learning} and Ren \etal~\cite{ren2016coarse} predicted edge maps with FCN-based model and inferred the layout with vanishing points or lines following the widely used proposal-ranking scheme. Dasgupta \etal~\cite{dasgupta2016delay} proposed a similar approach to estimate the surface label heatmaps instead of edge maps.
Zhao \etal~\cite{zhao2017physics} alternatively trained their model on large scale semantic segmentation dataset then transferred semantic features to edge maps.
RoomNet~\cite{lee2017roomnet} encoded the ordered room layout keypoint locations into 48 keypoint heatmaps for 11 room types.
Although this representation could be decoded easily, it required data with predefined room types and annotation of ordered corners for their keypoint heatmaps.
Unfortunately, none of the existing methods for single-view layout estimation have been designed for the layout of general room types. We solve this problem with our proposed flat layout representation and dynamic programming based post-processing which do not rely on the predefined layout topologies.

Several works targeted at estimating indoor layout for panoramic images.
Zou \etal~\cite{zou2018layoutnet} predicted the per-pixel corner probability map and boundary map like the cases in perspective images.
Yang \etal~\cite{yang2018dula} combined surface semantic with ceiling view and floor view.
Sun \etal~\cite{sun2019horizonnet} encoded boundaries and wall-wall existence in their "1D representation" to recover layouts from 360$^{\circ}$ panoramas. 

Sun \etal~\cite{sun2019horizonnet} is the most related method to ours. However, their representation was only suitable for 360$^{\circ}$ panoramic images, in which the ceiling-wall and floor-wall boundaries exist for every column under equirectangular projection. Their post-processing assume the layout formed a closed loop which is often true in panorama images but not the case for perspective image.

%% file: sec-approach.tex
\section{Approach} \label{sec:approach}

\subsection{Pre-processing} \label{sssec:approach_pre}
In the context of 360$^{\circ}$ panorama, aligning input image by three orthogonal vanishing points in the pre-processing phase is a common practice for layout estimation~\cite{zou2018layoutnet,fernandez2018panoroom,yang2018dula,sun2019horizonnet}.
On the other hand, existing works for perspective images layout estimation only use the vanishing point information in post-processing phase~\cite{ren2016coarse,dasgupta2016delay,zhao2017physics,zhang2019edge}.
To the best of our knowledge, we are the first exploiting vanishing point information in the pre-processing phase (before the deep model) for a single perspective image layout estimation task.

\begin{figure}
\centering
\begin{subfigure}{.5\linewidth}
  \centering
  \includegraphics[width=.9\linewidth]{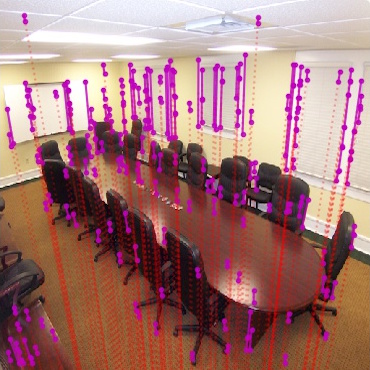}
  \caption{Source image}
  \label{fig:preproc_before}
\end{subfigure}%
\begin{subfigure}{.5\linewidth}
  \centering
  \includegraphics[width=.9\linewidth]{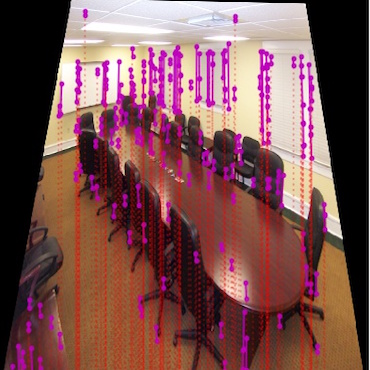}
  \caption{Rectified image}
  \label{fig:preproc_after}
\end{subfigure}
\caption{In pre-processing, we transform the source image (\ref{fig:preproc_before}) such that all wall-wall boundaries are parallel to image Y-axis (\ref{fig:preproc_after}). The magenta line segments are the inlier of the vertical vanishing point ($vp_Z$). The red dotted lines connect the center of line segments and $vp_Z$.}
\label{fig:preproc}
\end{figure}

To facilitate our flat room-layout representation (Sec.~\ref{ssec:flat_rep}), we want to rectify images such that all wall-wall boundaries are parallel to the image Y axis. This requirement can be easily achieved by detecting the vertical vanishing point ($vp_Z$) and constructing a homography to transform $vp_Z$ to infinite of the Y-axis of the image. To detect a $vp_Z$, we extract line segments using LSD~\cite{von2010lsd} and keep only line segments pointing vertically (pointing direction larger or smaller than $\pm  45^{\circ}$). The most voted point is detected by RANSAC as the $vp_Z$. Fig.~\ref{fig:preproc} depicts the effect of the pre-processing.

We apply the same pre-processing to all the training and testing images.

\subsection{Flat Room Layout Representation} \label{ssec:flat_rep}
\begin{figure}
   \centering
  \setlength\tabcolsep{1pt}
\includegraphics[width=0.8 \linewidth]{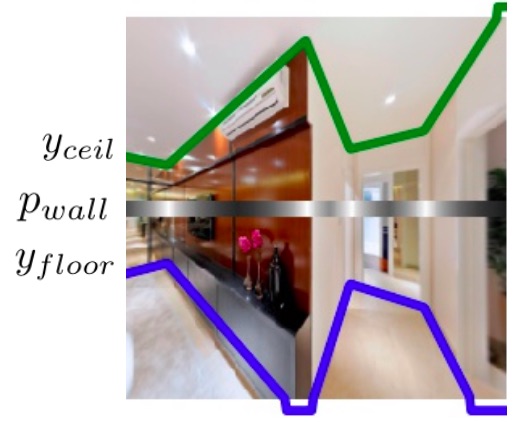}
    \caption{Visualization of our flat target output representation. $p_{wall}$ denotes the existence of wall-wall boundary. $y_{ceil}$ and $y_{floor}$ (plotted in green and blue) denote the position of ceiling-wall boundary and floor-wall boundary. Note that $y_{ceil}, y_{floor}$ and $p_{wall}$ are just three row vectors in effect.}
    \label{fig:representation}
\end{figure}

We introduce a flat target output representation of room layout that could be efficiently decoded to a layout. As illustrated in Fig.~\ref{fig:representation}, our flat output representation comprises three row vectors: $y_{ceil}$, $y_{floor}$, and $p_{wall}$, of the same length as image width, and two classifiers: $p_{ceil}$ and $p_{floor}$.

Each column of $y_{ceil}$ represents the position of the ceiling-wall boundary at the corresponding column of the image as a scalar. 
In the same way, $y_{floor}$ represents the position of the floor-wall boundary. 
$p_{wall}$ indicates the existence of wall-wall boundary at each column as a probability.
The two classifiers $p_{ceil}$ and $p_{floor}$ stand for whether the input image contains ceiling-wall boundary and floor-wall boundary or not. We explain the motivation of designing these two classifiers in Sec.~\ref{ssec:result_analy}.

The values of $y_{ceil}$ and $y_{floor}$ are normalized to $[0, 1]$. And for a column $i$ where ceiling-wall and floor-wall boundary do not exist, we set $y_{ceil}(i)$ to $-0.01$ and $y_{floor}(i)$ to $1.01$.
Considering that assigning $p_{wall}$ with 0/1 labels would result in a strongly class-imbalanced ground truth, we define $p_{wall}(i) = 0.96^{dx}$ where $i$ indicates the $i$th column and $dx$ is the distance from the $i$th column to the nearest wall-wall boundary.

\subsection{Network Architecture} 
\begin{figure*}
   \centering
   \setlength\tabcolsep{1pt}
\includegraphics[width=0.96 \linewidth]{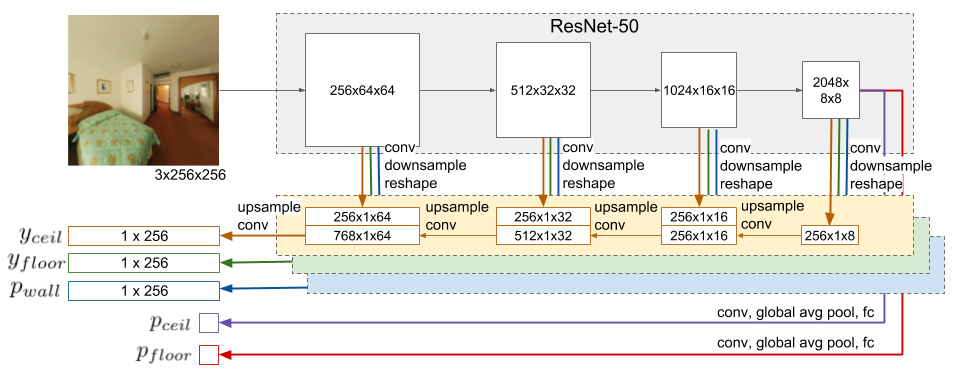}
    \caption{An illustration of the \modelname network architecture.}
    \label{fig:model_arch}
\end{figure*}

An overview of the \modelname network architecture is illustrated in Fig.~\ref{fig:model_arch}.
The \modelname network takes a single RGB image with the dimension $3 \times 256 \times 256$ (channel, height, width) as input.
Our network is built upon ResNet-50 \cite{HeZRS16}, followed by five output branches respectively for the five output targets defined in Sec.~\ref{ssec:flat_rep}. 

The two branches for classifiers $p_{ceil}, p_{floor}$ take the output feature maps from the last ResNet-50 block as input. Both branches consist of two convolution layers with $1 \times 1, 3 \times 3$ kernel sizes, a global average pooling layer which reduces the spatial dimension to 1x1, and one final fully-connected layer with 2 class softmax activation.

The structures of the three flat decoder branches for predicting $y_{ceil}, y_{floor}, p_{wall}$ are exactly the same without sharing weights.
We design the decoder for capturing both high-level global features and low-level local features.
Following the spirit of U-net~\cite{ronneberger2015u}, which gradually fuses features from deeper layers with features from shallower layers, we adopt a contracting-expanding structure, in which ResNet-50 serves as the contracting path (upper part in Fig.~\ref{fig:model_arch}) and the flat decoder branch is the expanding path (lower part of the figure).
More specifically, the ResNet-50 comprises four blocks, and each outputs feature maps with half spatial resolution compared to that of the previous block. 
To reduce the number of parameters, we add a sequence of convolution layers after each ResNet-50 block which reduce the number of channels and height by a factor of 4 and 8 respectively, then reshape the feature maps to height 1 to obtain flat feature maps.
Every step in the expanding path comprises an upsampling which doubles only the width of the flat feature maps followed by three convolutions with kernel size $1 \times 1, 1 \times 3, 1 \times 1$. 
At the last step, we upsample the width of flat feature maps to four times larger and apply two convolution layers to reduce the channel to 1, yielding the final output with the dimension $1 \times 256$ (image width). All the convolution layers except the last one are followed by ReLU and BatchNorm~\cite{batchnorm}.

\subsection{Post-processing} \label{ssec:post_proc}
Based on our flat representation, we propose an intuitive and efficient post-processing algorithm which is capable of generating layouts for general room layout topologies not limited to ``box-shape'' (namely, the images taken in cuboid-shaped rooms). An overview of our post-processing algorithm is: \emph{i}) peak finding to detect corners' x positions,  then \emph{ii}) special cases checking, finally \emph{iii}) using dynamic programming to determine the actual positions of all corners.

As a reminder, the $p_{wall}'$ predicted by our model is the probability of each image column being a wall-wall boundary. $y_{ceil}'$ and $y_{floor}'$ are the position of ceiling-wall and floor-wall boundary at each image column and any out of image plane position indicating no boundary at that image column. $p_{ceil}'$ and $p_{floor}'$ are the results from two softmax classifier branches telling us whether to ignore $y_{ceil}'$ or $y_{floor}'$.

\paragraph{Wall-wall positions:} We first extract information about the position of the wall-wall boundary from model estimated $p_{wall}'$. We use a smoothing filter with window size covering $5\%$ of the image width, then finding signal peaks with the same window size. Peaks with probability lower than $0.5$ are removed. The remain peaks telling the column position of all wall-wall boundaries which we denote as a set of image x positions $W_x$. \newline

For the remaining post-processing description, we will only explain using $y_{ceil}'$ and $W_x$. As we construct ceiling corners and floor corners independently with the same determined $W_x$, the same algorithm could be applied to $y_{floor}'$.

\paragraph{Two Special Cases:}
\emph{i}) $S_x$ is an empty set. If no wall-wall boundary is found in the image, we will simply predict a ceiling-wall boundary with linear regression on $y_{ceil}'$.
\emph{ii}) There are more than $99\%$ of columns of $y_{ceil}'$ are predicted to be out of image plane. Since the result suggesting that no ceiling-wall boundary in the image plane, we estimate the ceiling corners as $\{(x, 0) ~|~ x \in W_x\}$. (In the case of floor corners, the result is $\{(x, H-1) ~|~ x \in W_x\}$ where $H$ is image height.)

\begin{figure}
   \centering
  \setlength\tabcolsep{1pt}
\includegraphics[width=0.95 \linewidth]{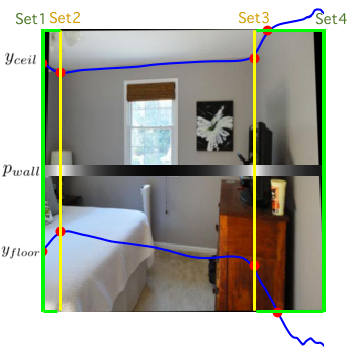}
    \caption{Illustration of the candidate points sets in our post-processing procedure.
    After finding $N$ peaks in the network predicted wall-wall existence, we generate $N+2$ candidate points sets, which include $N$ columns on the X coordinates of the $N$ peaks (yellow), the left-most edge, and the right-most edge (green).
    The corners decoded with the post-processing are plotted in red dots. Please see Sec.~\ref{ssec:post_proc} for detailed description of our post-processing algorithm.}
    \label{fig:post-process}
\end{figure}

\paragraph{Dynamic Programming:} With the detected $N = |W_x|$ peaks, we generate $N+2$ candidate points sets according to the positions of $x \in W_x$. Please see Fig.~\ref{fig:post-process} for better understanding. The generated candidate points sets are denoted as $S_L, S_1, S_2 \cdots, S_N, S_R$ where we will select a point in each set (red dots in the figure) as actual ceiling corners. $S_L$ and $S_R$ (green dots in the figure) are the sets of points all located at the image border with $x$ position less than $min(W_x)$ or greater than $max(W_x)$ respectively. The middle sets $S_1, \cdots, S_N$ (yellow dots in the figure) are $S_i = \{(W_x^{(i)}, y) ~|~ y \in {0, 1, \cdots, H-1}\}$ where $W_x^{(i)}$ is the $i$'th smallest element of $W_x$.

The raw estimated ceiling-wall boundary (blue dots in upper part of Fig.~\ref{fig:post-process}) is split also according to $W_x$, resulting in $N+1$ sets $P_0, P_1, \cdots, P_{N}$

To select a point in each set as the final actual ceiling corners (red dots in Fig.~\ref{fig:post-process}), we define loss as the average Euclidean distance of all the raw estimated ceiling-wall boundary positions (blue dots in the figure) to the estimated layout. More specifically, a point in $P_i$ producing a loss by the distance to the line connecting the selected two points in $S_i$ and $S_{i+1}$.
The loss function $V^{(j)}_i$ is defined to be minimum loss from $S_L$ to $S_i$ with $j$'th element of $S_i$ is selected.
To find the layout with minimum loss, we exploit dynamic programming. The recursive relationship is obvious and is provided in Eq.~\ref{eq:dp}
\begin{equation} \label{eq:dp}
\begin{split}
V^{(j)}_1 &= min_k ~ d(P_0, \overline{S_L^{(0)} S_1^{(k)}}) \,, \\
V^{(j)}_{i+1} &= min_k ~ V^{(k)}_i + d(P_i, \overline{ S_i^{(k)} S_{i+1}^{(j)} }) \,. \\
\end{split}
\end{equation}
The layout with minimum loss can be extracted by backtracking from $argmin_{p \in V_R} V_R(p)$ where $V_R = V_{N+1}$. The time complexity of the overall algorithm is $O(N \cdot H^2)$ and takes roughly 400ms for a frame. There is still much room for speedup as our implementation is based on Python and many redundant candidates points in each set can be removed in a heuristic manner.

\subsection{Reconstruct Layout Piece-wise Planar in 3D} 
To reconstruct the recognized layout in 3D, we make a few assumptions: (\emph{i}) the pre-processing correctly transform the $vp_Z$ to infinite Y axis of image, (\emph{ii}) the floor and ceiling are planes orthogonal to gravity direction, and (\emph{iii}) the distance between camera to floor and camera to ceiling are both 1 meter.

In below explanation, $x, y$ are used to indicate the pixel position on the image plane where the origin is defined as the image center, and the right and bottom are defined as positive directions of $x$ and $y$ respectively. $X, Y, Z$ are used to indicate the position in 3D where the camera center is located at $(0, 0, 0)$ and floor plane is $Z=-1$.

Before further reconstruction, we have to infer the imaginary pixel distance, $f$, between camera center and image plane. We use assumption (\emph{i}) and three floor corners $(x_0, y_0), (x_1, y_1), (x_2, y_2)$ in the image which are considered to form an right angle in 3D with $(x_0, y_0)$ on the angle. The $f$ is then can be solved by Eq.~\ref{eq:f} (See supplementary for detail derivation).
\begin{equation} \label{eq:f}
f^2 = \frac{ x_1 x_2 y_0^2 + x_0^2 y_1 y_2 - x_0 x_1 y_0 y_2 - x_0 x_2 y_0 y_1 }{ y_0 y_1 + y_0 y_2 - y_0^2 - y_1 y_2 } \,.
\end{equation}
With the $f$ and a given $Z$ we get equations $X = \frac{x Z}{y}$ and $Y = \frac{f Z}{y}$ for mapping from image coordinate to world coordinate under our notation. Combined with assumptions (\emph{ii}) and (\emph{iii}), we can obtain the world coordinate of each corners on floor and ceiling for texture mapping in 3D viewer. Some reconstructed results are in Fig.~\ref{fig:teaser} and Fig.~\ref{fig:qual_general3d}.

%% file: sec-experiments.tex
\section{Experiments}

\subsection{Implementation Details}
All input RGB images and ground truth are resized to 256 $\times$ 256.
The L2 loss is used for the ceiling-wall boundary ($y_{ceil}$) and floor-wall boundary ($y_{floor}$), while the $y_{ceil}$ loss is not computed if the model output is already smaller than 0 when the ground truth is -0.01 in the columns where ceiling-wall boundary does not exist, and likewise, the $y_{floor}$ loss is not counted if the model output is greater than 1 when the ground truth is 1.01.
We adopt the binary cross-entropy loss for wall-wall boundary ($p_{wall}$) and use the cross-entropy loss for both ceiling-wall boundary classifier ($p_{ceil}$) and floor-wall boundary classifier ($p_{floor}$).
The Adam optimizer~\cite{adam} is employed to train the network for 150 epochs with batch size 16 and learning rate 0.0002.
Similar to \cite{liu2015parsenet,chen2017rethinking}, we employ a ”poly” learning rate policy, where the initial learning rate is multiply by $(1-iter_{now}/iter_{max})^{0.9}$.

\subsection{Evaluation Details}
We use two standard quantitative evaluation metric for room layout estimation. \emph{i}) Pixel Error (PE) calculates the accuracy of per-pixel surface label between ground truth and estimation. \emph{ii}) Corner Error (CE) is defined by Euclidean distance between ground truth corners and estimated corners normalized by image diagonal length. LSUN~\cite{yu15lsun} official evaluation codes take the minimum CE among all possible matching and penalize $0.3$ for each extra or missing corners from ground truth.

In the pre-processing phase of our approach pipeline, images are rectified by homography (see Fig.~\ref{fig:preproc}) so the estimated corners or surface semantic can not be evaluated directly with the original ground truth. For a fairness comparison with literature, we project corners estimated by our model back to the original image and also rescale them to the original image resolution.

\subsection{Results on Hedau Dataset}
Hedau~\cite{hedau2009recovering} dataset consists of 209 training instances and 105 testing instances. We skip the Hedau training set and evaluate our approach trained on LSUN~\cite{yu15lsun} dataset directly on Hedau testing set. As the ground truth corners are not provided and also being consistent with the literature, we only use pixel error (PE) as the evaluation metric. The quantitative results on Hedau testing set compared with other methods are summarized in Table.~\ref{table:quan_hedau}. Our approach achieves state-of-the-art performance. Some qualitative results are provided in supplementary.

\begingroup
\renewcommand{\arraystretch}{1.1}
\begin{table}[h]
    \centering
    \begin{tabular}{l|c} 
        \hline
        Method & PE (\%) \\
        \hline\hline
        Hedau \etal (2009)~\cite{hedau2009recovering} & 21.20 \\ 
        Del Pero \etal (2012)~\cite{del2012bayesian} & 16.30 \\
        Gupta \etal (2010)~\cite{gupta2010estimating} & 16.20 \\
        Zhao \etal (2013)~\cite{zhao2013scene} & 14.50 \\
        Ramalingam \etal (2013)~\cite{ramalingam2013manhattan} & 13.34 \\
        Mallya \etal (2015)~\cite{mallya2015learning} & 12.83 \\
        Schwing \etal (2012)~\cite{schwing2012efficient} & 12.80 \\
        Del Pero \etal (2013)~\cite{del2013understanding} & 12.70 \\
        Izadinia \etal (2017)~\cite{izadinia2017im2cad} & 10.15 \\
        Dasgupta \etal (2016)~\cite{dasgupta2016delay} & 9.73 \\
        Zou \etal (2018)~\cite{zou2018layoutnet} & 9.69 \\
        Ren \etal (2016)~\cite{ren2016coarse} & 8.67 \\
        Lee \etal (2017)~\cite{lee2017roomnet} & 8.36 \\
        \hline
        \textbf{ours} & \textbf{5.01} \\
        \hline
    \end{tabular}

    \caption{
    Quantitative results on Hedau~\cite{hedau2009recovering} testing set.
    }
    \label{table:quan_hedau}
\end{table}
\endgroup

\subsection{Results on LSUN Dataset}
LSUN~\cite{yu15lsun} dataset consists of 4000 training instances, 394 validation instances, and 1000 testing instances. Because the ground truth of testing set is not available, we evaluate and compare with other approaches only on the validation set (also the case of~\cite{ren2016coarse,lee2017roomnet,zou2018layoutnet} where only result on validation set are reported). The quantitative results are shown in Table.~\ref{table:quan_lsun} where we achieve state-of-the-art performance. Some qualitative results are provided in supplementary.

\begingroup
\renewcommand{\arraystretch}{1.1}
\begin{table}[h]
    \centering
    \begin{tabular}{l|c|c} 
        \hline
        Method & CE (\%) & PE (\%) \\
        \hline\hline
        Hedau \etal (2009)~\cite{hedau2009recovering} & 15.48 & 24.23 \\ 
        Mallya \etal (2015)~\cite{mallya2015learning} & 11.02 & 16.71 \\
        Dasgupta \etal (2016)~\cite{dasgupta2016delay} & 8.20 & 10.63 \\
        Ren \etal (2016)~\cite{ren2016coarse} & 7.95 & 9.31 \\
        Zou \etal (2018)~\cite{zou2018layoutnet} & 7.63 & 11.96 \\
        Lee \etal (2017)~\cite{lee2017roomnet} & 6.30 & 9.86 \\
        \hline
        \textbf{ours} & \textbf{4.92} & \textbf{6.68} \\ 
        \hline
    \end{tabular}

    \caption{
    Quantitative results on LSUN testing set~\cite{yu15lsun}.
    }
    \label{table:quan_lsun}
\end{table}
\endgroup

\subsection{Results Analysis} \label{ssec:result_analy}
\paragraph{Data imbalance problem in LSUN:} We design the two classifier branch described in  Sec.~\ref{ssec:flat_rep} to suppress the false positive boundary regression (in other words, our model find a boundary while the ground truth is empty). The intuition is by observing the severe room type imbalance in LSUN datasets~\cite{yu15lsun}. We depicted the number of training instances according to the room layout types defined by LSUN in Fig.~\ref{fig:lsun_distribution}. The detailed definition of each room type is provided in supplementary. The number of instances belonging to type 2, 3, 7, 8, 10 which do not have floor-wall boundary only accounts for $1.9\%$ of the total number of training instances. This data imbalance problem makes our learning-based model tend always to predict a floor-wall boundary.

To further prove that the design of the two classifiers could ease the bias caused by type imbalance, we show the corner error with and without the help of the classifiers in Table.~\ref{table:abla_clf}. The results show that the classifier branches could help in the case that the floor-wall boundary or ceiling-wall boundary is outside the image plane.

\begin{figure}
   \centering
  \setlength\tabcolsep{1pt}
\includegraphics[width=1.0 \linewidth]{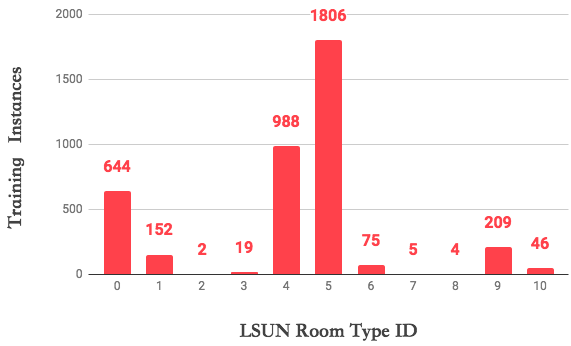}
    \caption{Distribution of the training samples according to the room layout types.}
    \label{fig:lsun_distribution}
\end{figure}

\begingroup
\renewcommand{\arraystretch}{1.1}
\begin{table}[h]
    \centering
    \begin{tabular}{c|c|c|c} 
        \hline
        \makecell{Types\\ID} & \makecell{No classifier\\CE(\%)} & \makecell{With classifier\\CE(\%)} & \makecell{Improvement\\CE(\%)} \\
        \hline\hline
        \multicolumn{4}{c}{Both floor-wall and ceiling-wall appear} \\
        \hline
        0 & 4.54 & 4.75 & -0.22 \\
        5 & 2.83 & 3.09 & -0.27 \\
        6 & 8.27 & 8.27 & 0.00 \\

        \hline\hline
        \multicolumn{4}{c}{Only floor-wall appear} \\
        \hline
        1 & 12.71 & 12.70 & +0.01 \\
        4 &  6.41 &  5.18 & \textbf{+1.23} \\
        9 &  9.86 &  8.65 & \textbf{+1.21} \\

        \hline\hline
        \multicolumn{4}{c}{Only ceiling-wall appear} \\
        \hline
        2 & - & - & - \\
        3 & 19.28 & 16.30 & \textbf{+2.98} \\
        8 & - & - & - \\

        \hline\hline
        \multicolumn{4}{c}{No floor-wall or ceiling-wall appear} \\
        \hline
        7 & - & - & - \\
        10 & 23.05 & 19.95 & \textbf{+3.10} \\

        \hline
    \end{tabular}

    \caption{Compare the corner error (CE) with and without the classifier branch in ablation manner on the validation set of LSUN dataset. Type 2, 8 and 7 are left empty because these types do not exist in the validation set.}
    \label{table:abla_clf}
\end{table}
\endgroup

\paragraph{Compare with Zhao \etal~\cite{zhao2017physics}:} An "alternative" method for room layout estimation is proposed by Zhao \etal~\cite{zhao2017physics} where they transferred information from larger scale semantic segmentation dataset (SUNRGBD~\cite{song2015sun}) by a complex training protocol while achieving outstanding performance, even the most recent state-of-the-art layout estimation approaches (\eg~\cite{lee2017roomnet,zou2018layoutnet}) did not outperform their results.
Comparing to our method, they achieve better result on LSUN dataset ($3.95 \%$ CE and $5.48 \%$ PE \vs ours $4.92 \%$ CE, $6.68 \%$ PE) while getting worse result on Hedau dataset than ours ($6.60 \%$ PE \vs ours $5.01 \%$ PE).
However, like other existing methods, they have to define a prior set of room type topologies (11 types in LSUN which also covers all possible types in Hedau). Besides, time complexity and the number of parameters of their post optimization algorithm increase linear to the number of room type. Our method, on the other hand, can extend to general room type without modifying our model and post-processing algorithm as we will show in Sec.~\ref{ssec:general}.

\subsection{General Layout Topology} \label{ssec:general}
\paragraph{Motivation:} Existing datasets for layout estimation from perspective images, i.e. LSUN~\cite{yu15lsun} and Hedau~\cite{hedau2009recovering} dataset, have the prior that at most two wall-wall boundaries are presented in the image. More specifically, LSUN dataset defined 11 layout types, and both LSUN and Hedau datasets considered only 5 categories of surface categories which are ceiling, floor, left wall, front wall, and right wall.

Existing methods for room layout estimation only take the layout topologies defined in LSUN dataset into consideration. They require further modification to generate layout which is not defined in LSUN. For instance, RoomNet~\cite{lee2017roomnet} need to define all layout topologies for their model. Hypotheses ranking based algorithms~\cite{ren2016coarse,dasgupta2016delay,izadinia2017im2cad} have to design additional rules for generating the proposal for covering possible layout types.

Our approach, on the other hand, can handle general room layout topologies directly without modifying our model. Our post-processing for decoding the predicted flat room layout representation can be directly applied to produce general room layout.

\paragraph{Experiment:}  To verify the idea, we build a benchmark with the 40 panoramic images annotated as general room layout by HorizonNet~\cite{sun2019horizonnet}. We split the 40 rooms into 10 folds for 10-fold cross-validation.
If a room is in validation subsample, we capture 6 perspective images by uniformly rotate the camera within the panorama and remove instances without any boundaries or corners inside the image plane (where the camera is too close to the wall). For a room in training subsample, we capture the perspective data like the case of validation but also with Pano Stretch Augmentation proposed by HorizonNet~\cite{sun2019horizonnet}.
Fig.~\ref{fig:representation} show one of the captured perspective image.

For each validation subsample, we use the model trained on LSUN dataset and finetune on the other training subsamples with 25 epochs.

\paragraph{Results:} Each of the 40 rooms is validated once during the course of 10-fold cross-validation. We cluster the corner error (CE) by the number of wall-wall visible in the perspective image and show the result in Table.~\ref{table:quan_general}. We also show the result of the model trained only on LSUN dataset (which contains only 0, 1 and 2 number of wall-wall) in the third column. We observe that the model only trained on LSUN get severely degraded result on cases of 3 or more wall-wall boundaries. The results show that our approach is applicable to general layout topology with available training data. We show some qualitative 3d reconstructed layout in Fig.~\ref{fig:teaser} and Fig.~\ref{fig:qual_general3d}. More qualitative results are provided in supplementary.

\begingroup
\renewcommand{\arraystretch}{1.1}
\begin{table}[h]
    \centering
    \begin{tabular}{c|c||c|c} 
        \hline
        \makecell{Number of\\wall-wall} & \makecell{Number of\\instances} & \makecell{No finetune\\CE (\%)} & \makecell{Finetune\\CE (\%)} \\
        \hline\hline
        0 & 5 & 8.30 & 4.69 \\
        1 & 86 & 5.63 & 4.81 \\
        2 & 59 & 3.55 & 3.45 \\
        3 & 48 & 13.59 & 6.78 \\
        4 & 29 & 14.65 & 6.26 \\
        5 & 7 & 16.60 & 10.89 \\
        6 & 2 & 20.98 & 15.25 \\
        \hline\hline
         & 236 & 8.35 & 5.31 \\
        \hline
    \end{tabular}

    \caption{
    Results on general room layout benchmark constructed by us. The third column are testing directly with our LSUN pre-trained model while the fourth column is the result of finetuning in cross-validation manner. The corner error (CE) shown in the last row are averaged across all the instances. See Sec.~\ref{ssec:general} for more detail.
    }
    \label{table:quan_general}
\end{table}
\endgroup

\begin{figure*}
   \centering
   \setlength\tabcolsep{0pt}

\begin{tabular}{cccc}

(a) & (b) & (c) & (d) \\

\makecell{\includegraphics[width=0.245\linewidth]{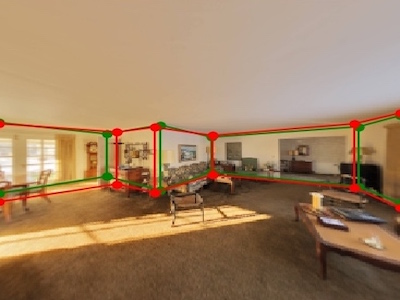}}&
\makecell{\includegraphics[width=0.245\linewidth]{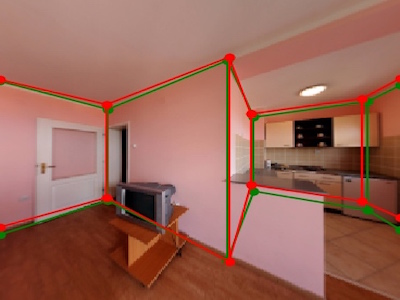}}&
\makecell{\includegraphics[width=0.245\linewidth]{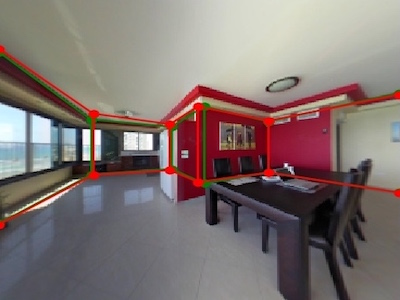}}&
\makecell{\includegraphics[width=0.245\linewidth]{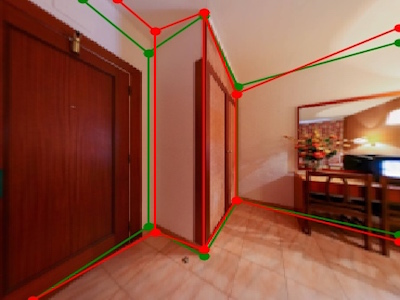}}
\\

\makecell{\includegraphics[width=0.25\linewidth]{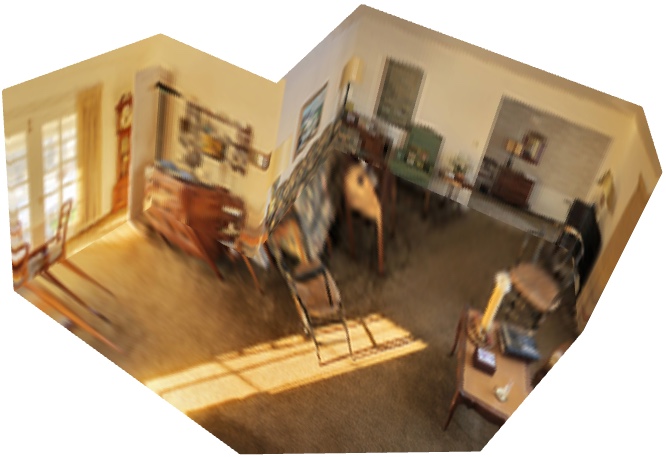}}&
\makecell{\includegraphics[width=0.25\linewidth]{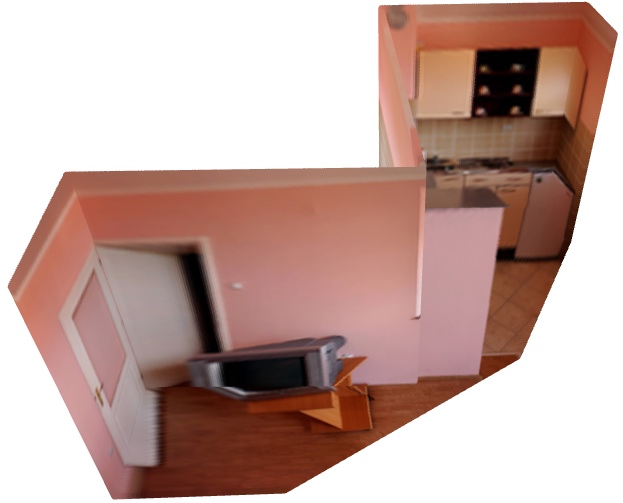}}&
\makecell{\includegraphics[width=0.25\linewidth]{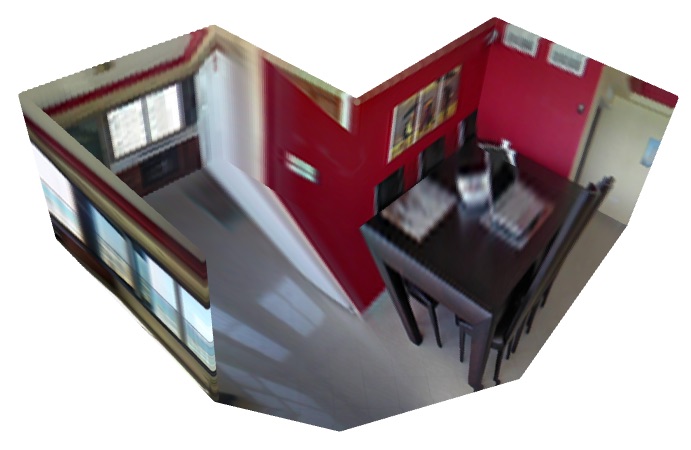}}&
\makecell{\includegraphics[width=0.25\linewidth]{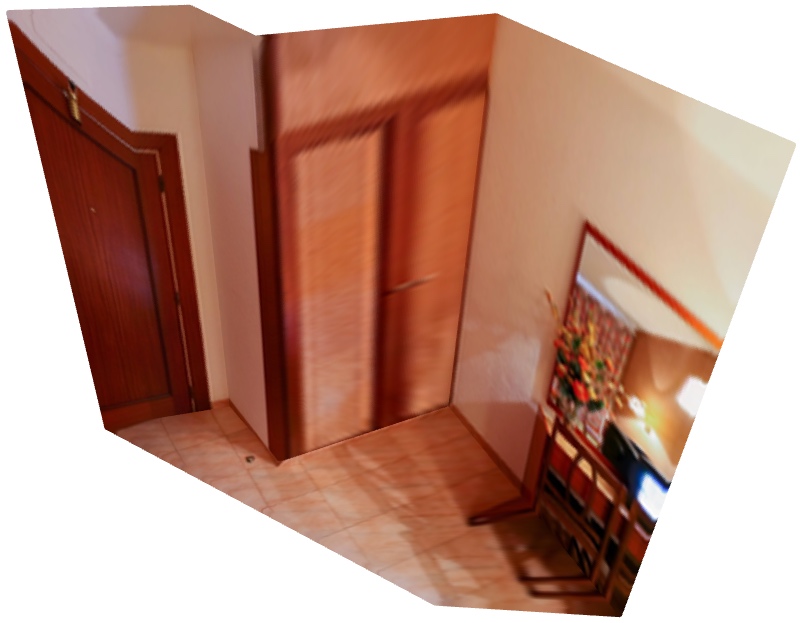}}
\\

\\

(e) & (f) & (g) & (h) \\
\makecell{\includegraphics[width=0.245\linewidth]{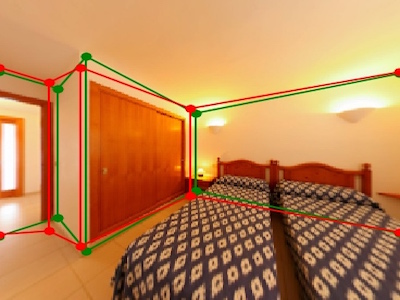}}&
\makecell{\includegraphics[width=0.245\linewidth]{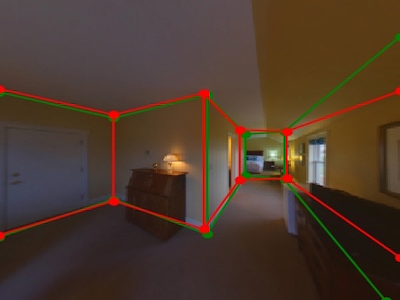}}&
\makecell{\includegraphics[width=0.245\linewidth]{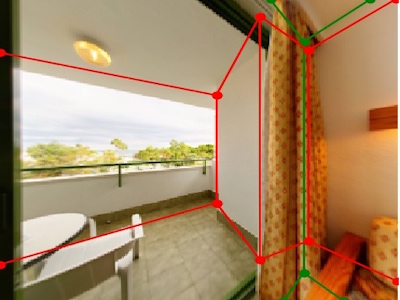}}&
\makecell{\includegraphics[width=0.245\linewidth]{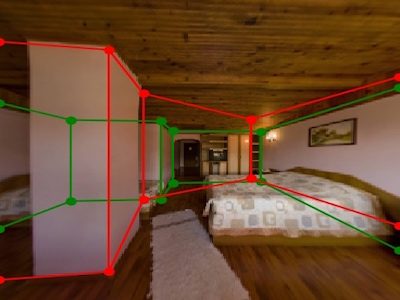}}
\\

\makecell{\includegraphics[width=0.25\linewidth]{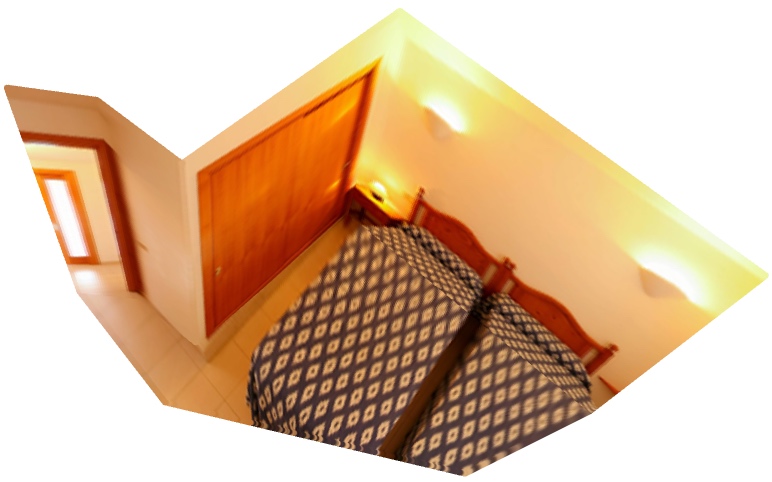}}&
\makecell{\includegraphics[width=0.25\linewidth]{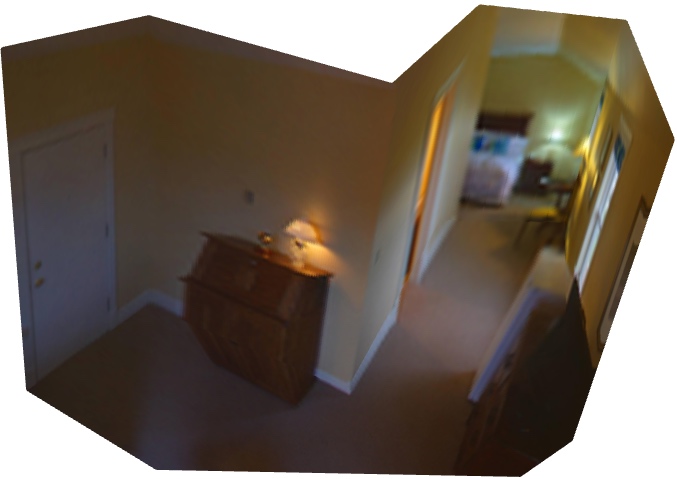}}&
\makecell{\includegraphics[width=0.25\linewidth]{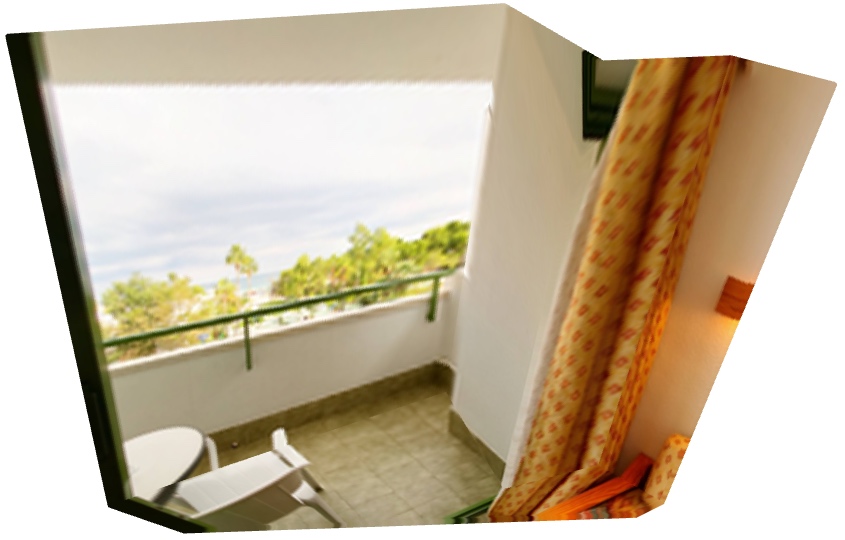}}&
\makecell{\includegraphics[width=0.25\linewidth]{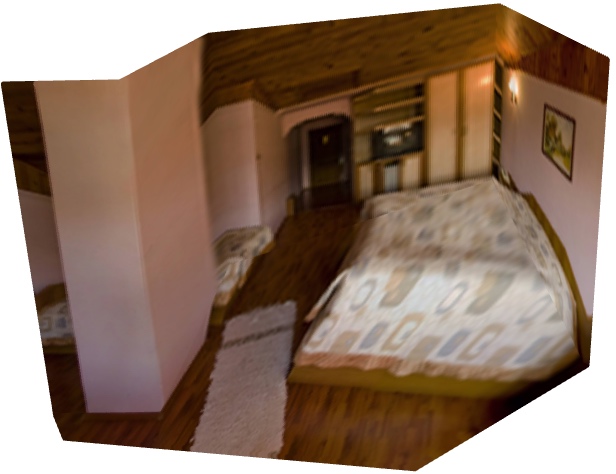}}
\\

\end{tabular}
    
    \caption{Some qualitative results of 3d reconstructed layout by our approach. The green dots and lines are ground truth and the red ones are estimated by our approach. Please see Sec.~\ref{sec:approach} for more detail about our method. In sample (g), our model failed to fit the true indoor layout but surprisingly recognize the shape of balcony thus producing visually acceptable result. Sample (h) is a challenging example which our approach fail and recognize the beam column as wall.}
    \label{fig:qual_general3d}
\end{figure*}

%% file: sec-conclusion.tex
\section{Conclusion}

We have presented a new approach, \modelname, which is able to recover the layout of general indoor room types from a single RGB image using a flat target representation. The proposed deep model is trained to estimate layout under the flat representation. To extract layout from flat representation, we exploit dynamic programming as post-processing which is fast and effective. Our approach achieves state-of-the-art performance on existing room layout datasets. Besides, we quantitatively and qualitatively show that our approach can also work directly on general room topology by only changing the training data, which overcomes the box-shape limitation of existing datasets and methods.

%% file: supp_lsun_types.tex
\section{Definition of LSUN Room Types}
\begin{figure}[!htb]
  \centering
  \setlength\tabcolsep{1pt}
\includegraphics[width=1\linewidth]{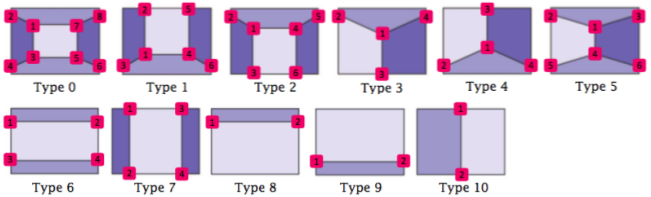}
    \caption{The 11 room types defined in the LSUN dataset.}
    \label{fig:lsun_types}
\end{figure}

%% file: supp_derivation_3d.tex
\section{Derivation for 3D Reconstruction}
We use $Y$ and $Z$ for front-back and up-down direction respectively in world coordinate system for our convenient, so $X = \frac{xZ}{y}$ and $Y = \frac{fZ}{y}$ are obtained by rearranging $x = \frac{fX}{Y}$ and $y = \frac{fZ}{Y}$.
For later 3D reconstruction, we need to solve the unknown term $f$ first, which can be done by using three points $(x_0, y_0), (x_1, y_1), (x_2, y_2)$ estimated by our model on the image plane. The three points are considered to be on the floor and forming $90^{\circ}$ angle with $(x_0, y_0)$ on the vertex.
As $Z_0 = Z_1 = Z_2 = -1$ in our assumption, we have
\begin{equation} 
\label{eqn:sup_Z}
(X_1 - X_0)(X_2 - X_0) + (Y_1 - Y_0)(Y_2 - Y_0) = 0 \,.
\end{equation}
By expanding (\refeq{eqn:sup_Z}), we have a linear equation for $f^2$:
\begin{equation} \label{eqn:sup_f}
\left(x_1 y_0 - x_0 y_1 \right)
\left(x_2 y_0 - x_0 y_2 \right) +
\left( y_0-y_1 \right)
\left( y_0-y_2 \right) f^2 = 0 \,.
\end{equation}
Based on the solution of $f$, one of the wall-wall intersection is guaranteed to be orthogonal.

%% file: supp_qual_hedau.tex
\section{Qualitative Results on Hedau Dataset}

\begin{figure}[H]
   \centering
   \setlength\tabcolsep{0pt}

\begin{tabular}{ccc}

(a) & 
\makecell{\includegraphics[width=0.44\linewidth,height=0.33\linewidth]{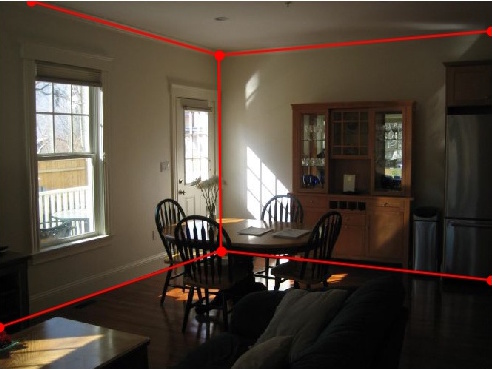}}&
\makecell{\includegraphics[width=0.44\linewidth,height=0.33\linewidth]{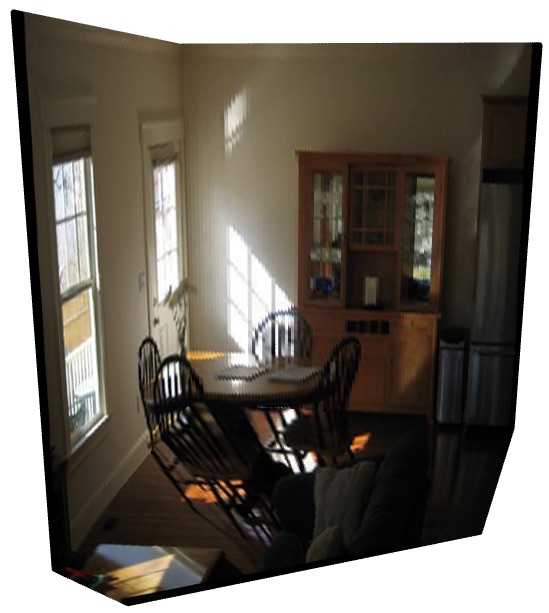}}
\\
(b) & 
\makecell{\includegraphics[width=0.44\linewidth,height=0.33\linewidth]{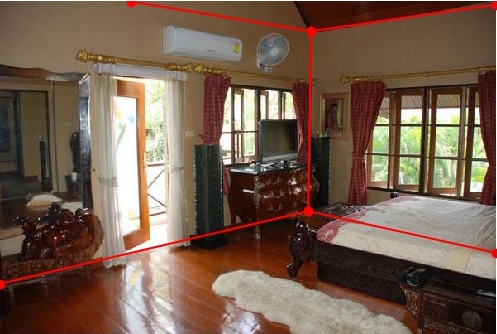}}&
\makecell{\includegraphics[width=0.44\linewidth,height=0.33\linewidth]{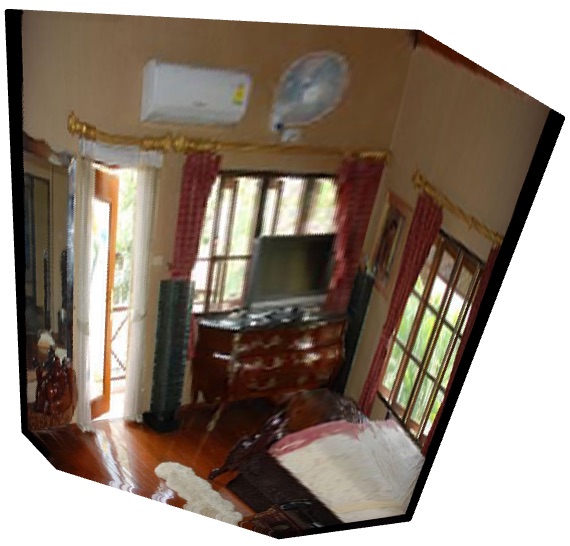}}
\\
(c) & 
\makecell{\includegraphics[width=0.44\linewidth,height=0.33\linewidth]{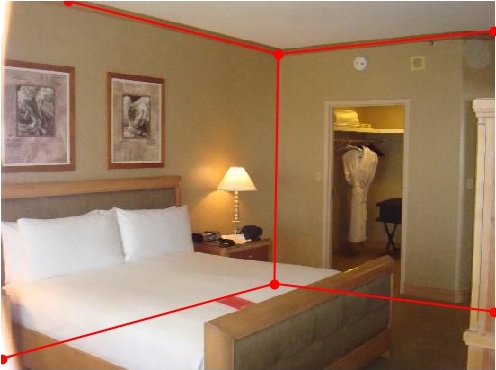}}&
\makecell{\includegraphics[width=0.44\linewidth,height=0.33\linewidth]{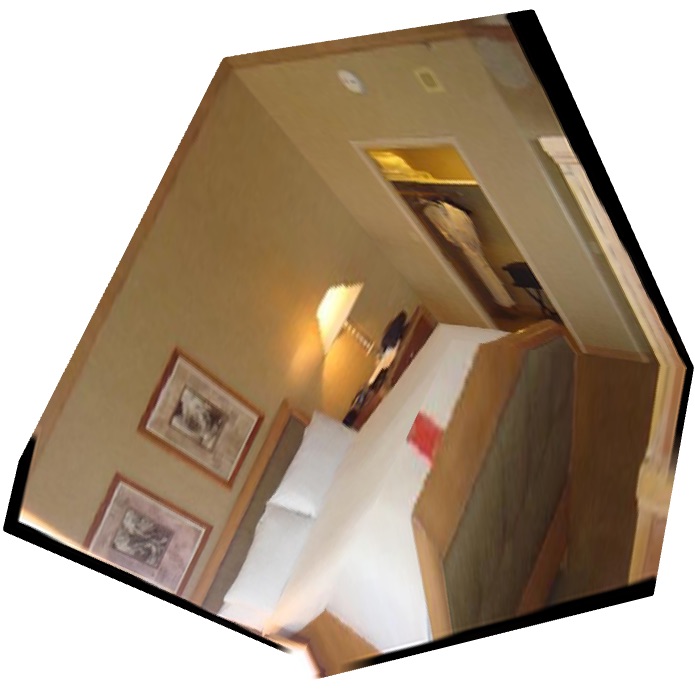}}
\\
(d) & 
\makecell{\includegraphics[width=0.44\linewidth,height=0.33\linewidth]{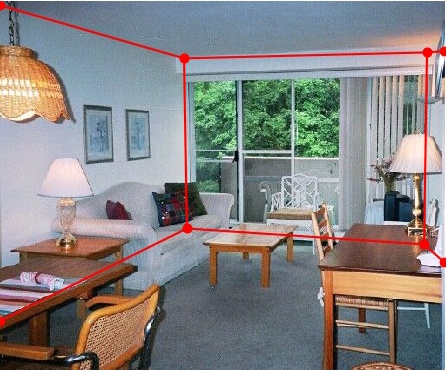}}&
\makecell{\includegraphics[width=0.44\linewidth,height=0.33\linewidth]{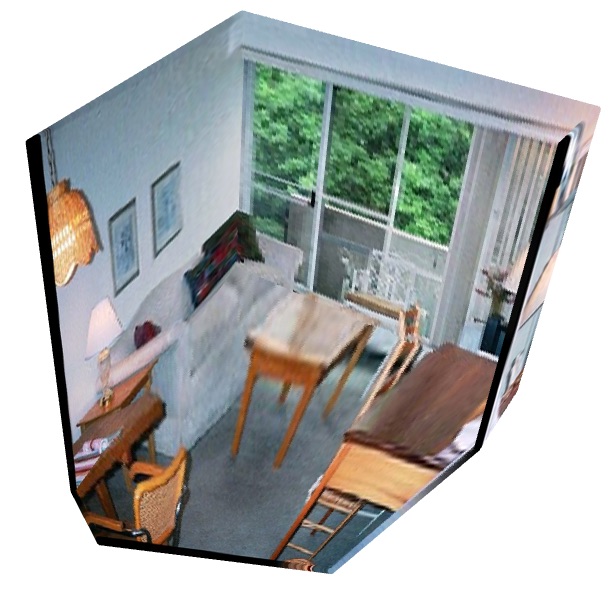}}
\end{tabular}

    \caption{Qualitative results on the Hedau test set. The results are separately sampled from four groups that comprise the predictions with the best 0--25\%, 25--50\%, 50--75\% and 75--100\% pixel errors (displayed from the first to the fourth row). The red lines depict the estimated layout.}
    \label{fig:qual_hedau}
\end{figure}

%% file: supp_qual_lsun.tex
\section{Qualitative Results on LSUN Dataset}

\begin{figure*}[htb]
   \centering
   \setlength\tabcolsep{1pt}

\begin{tabular}{cccc}

(a) & (b) & (c) & (d) \\

\makecell{\includegraphics[width=0.24\linewidth,height=0.16\linewidth]{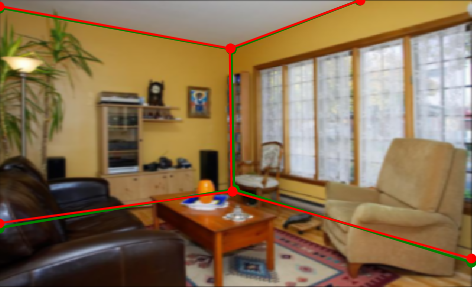}}&
\makecell{\includegraphics[width=0.24\linewidth,height=0.16\linewidth]{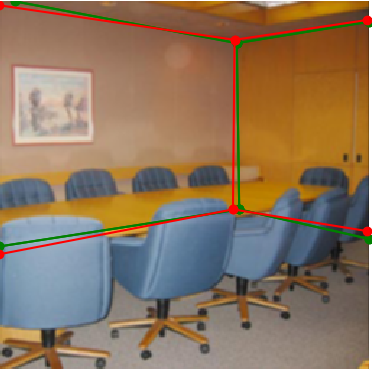}}&
\makecell{\includegraphics[width=0.24\linewidth,height=0.16\linewidth]{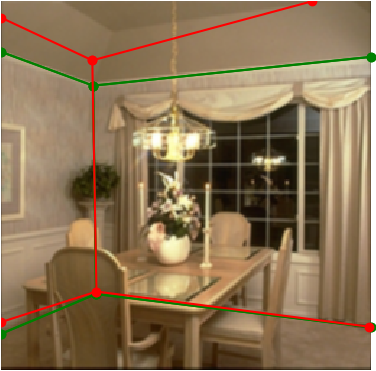}}&
\makecell{\includegraphics[width=0.24\linewidth,height=0.16\linewidth]{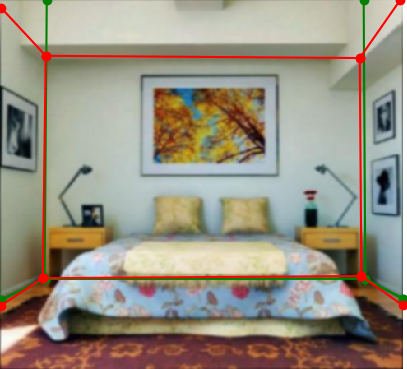}}
\\
\makecell{\includegraphics[width=0.24\linewidth,height=0.16\linewidth]{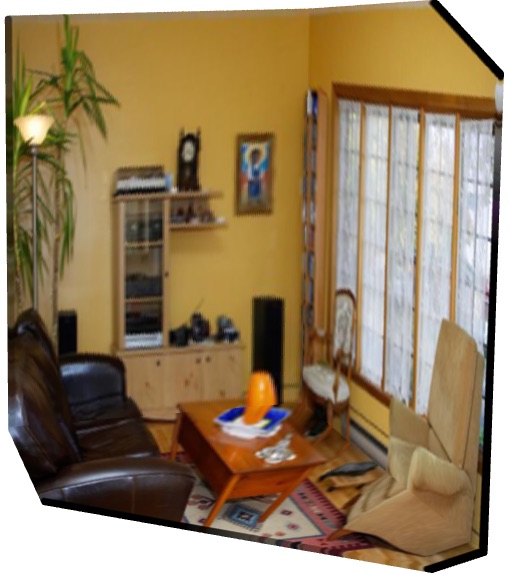}}&
\makecell{\includegraphics[width=0.24\linewidth,height=0.16\linewidth]{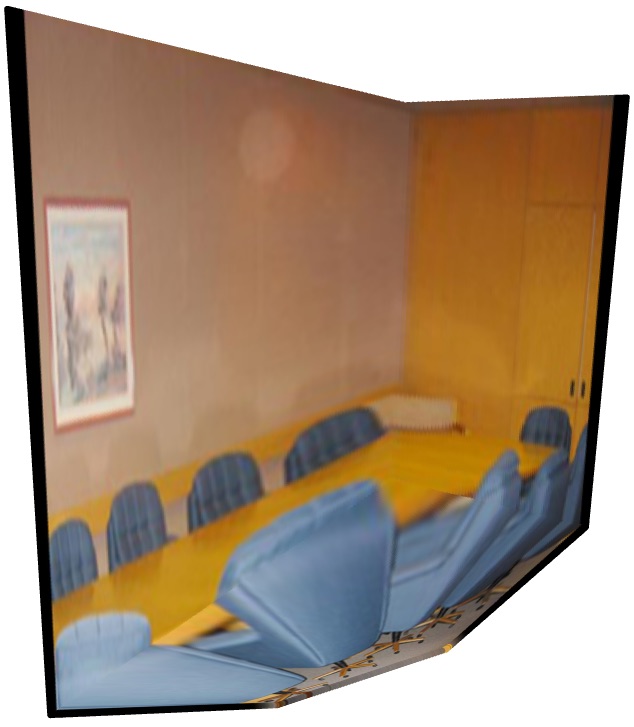}}&
\makecell{\includegraphics[width=0.24\linewidth,height=0.16\linewidth]{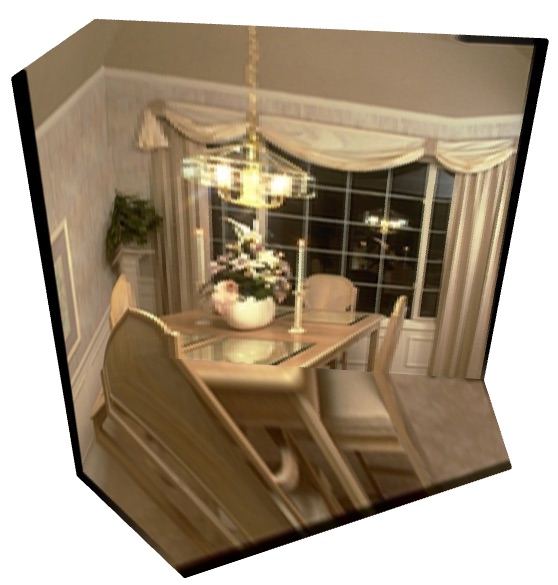}}&
\makecell{\includegraphics[width=0.24\linewidth,height=0.16\linewidth]{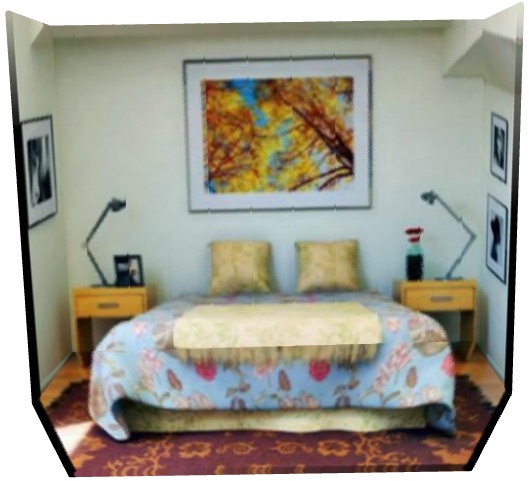}}

\\
\\
\makecell{\includegraphics[width=0.24\linewidth,height=0.16\linewidth]{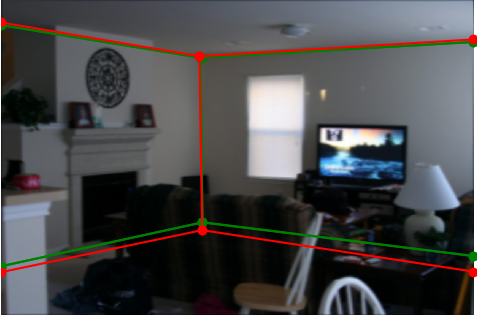}}&
\makecell{\includegraphics[width=0.24\linewidth,height=0.16\linewidth]{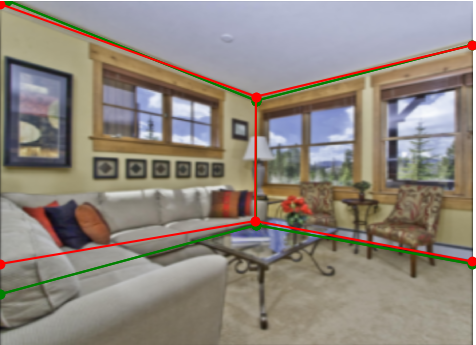}}&
\makecell{\includegraphics[width=0.24\linewidth,height=0.16\linewidth]{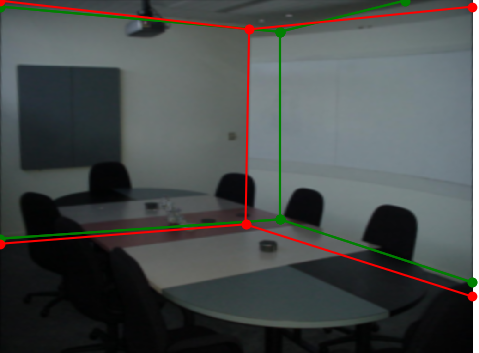}}&
\makecell{\includegraphics[width=0.24\linewidth,height=0.16\linewidth]{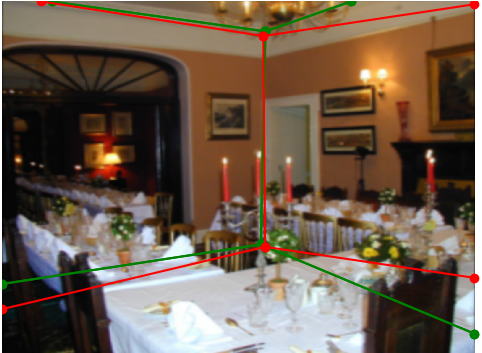}}
\\
\makecell{\includegraphics[width=0.24\linewidth,height=0.16\linewidth]{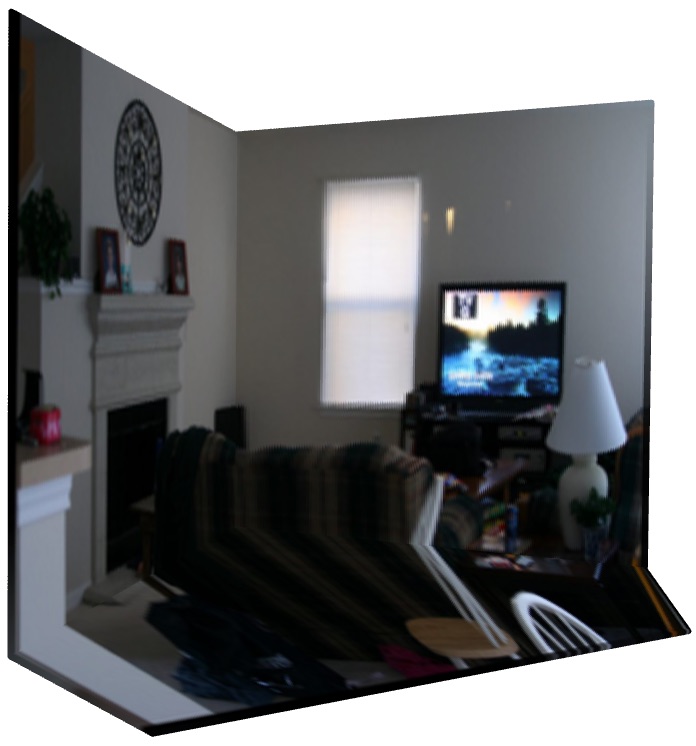}}&
\makecell{\includegraphics[width=0.24\linewidth,height=0.16\linewidth]{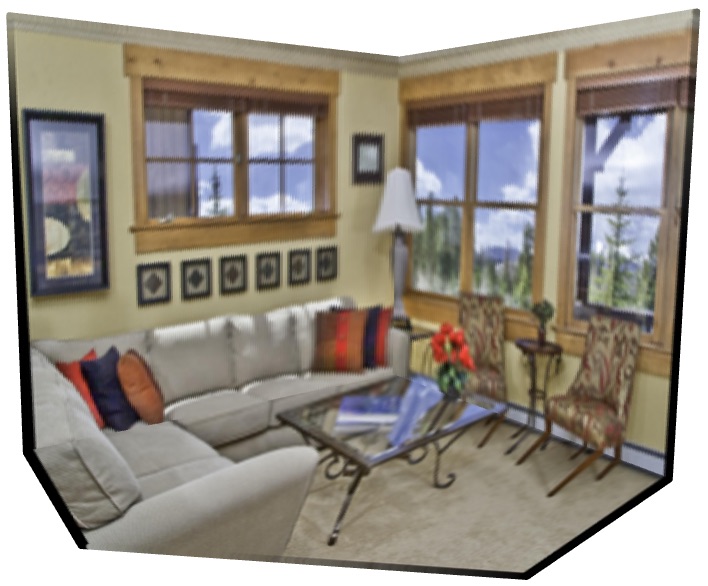}}&
\makecell{\includegraphics[width=0.24\linewidth,height=0.16\linewidth]{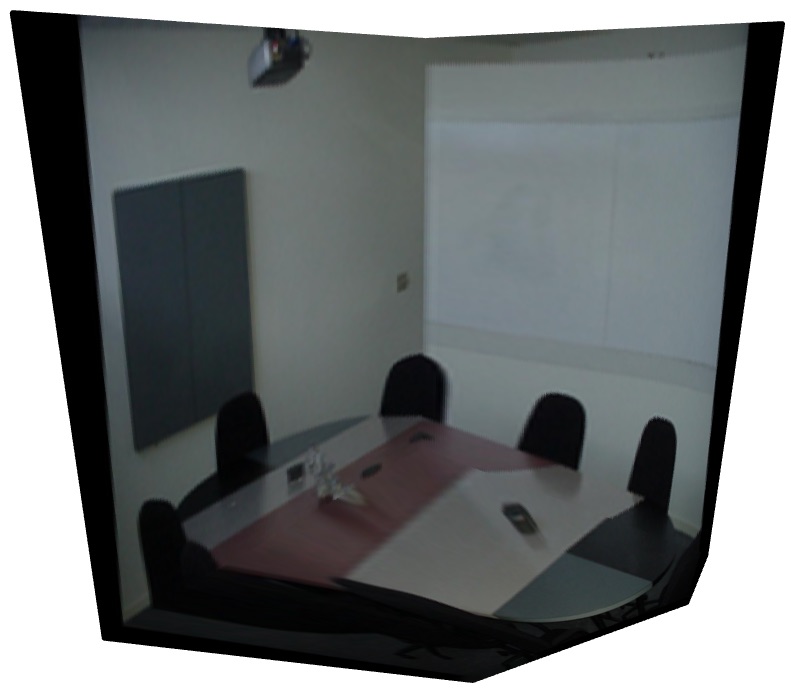}}&
\makecell{\includegraphics[width=0.24\linewidth,height=0.16\linewidth]{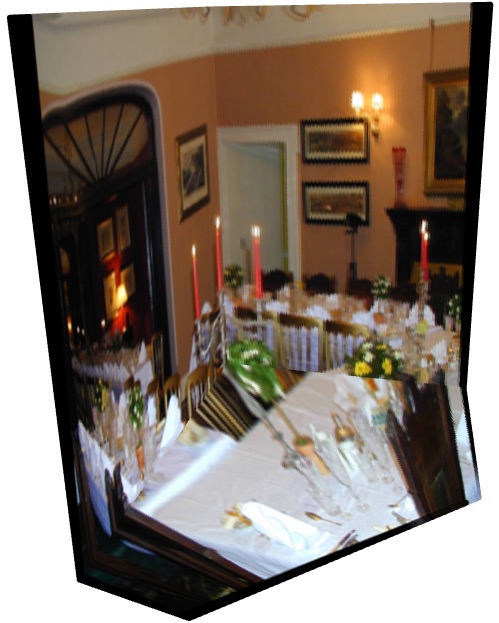}}

\\
\\
\makecell{\includegraphics[width=0.24\linewidth,height=0.16\linewidth]{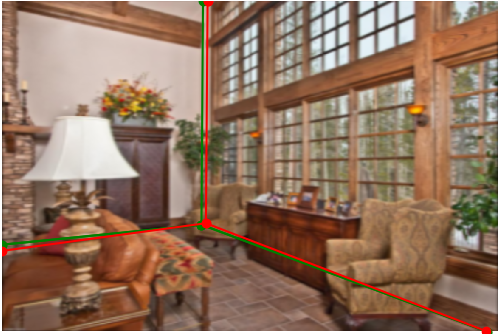}}&
\makecell{\includegraphics[width=0.24\linewidth,height=0.16\linewidth]{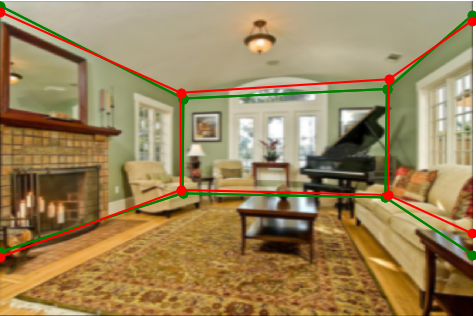}}&
\makecell{\includegraphics[width=0.24\linewidth,height=0.16\linewidth]{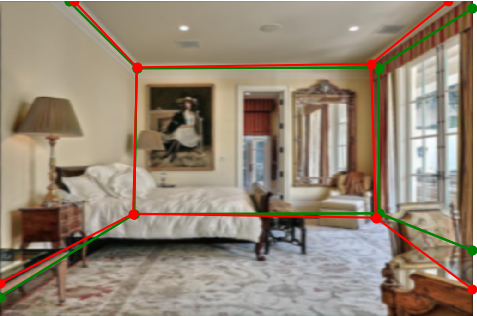}}&
\makecell{\includegraphics[width=0.24\linewidth,height=0.16\linewidth]{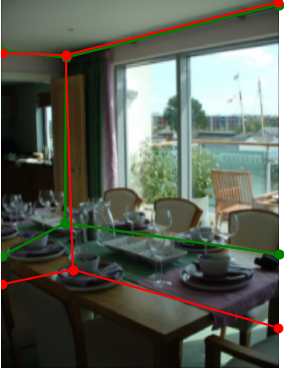}}
\\
\makecell{\includegraphics[width=0.24\linewidth,height=0.16\linewidth]{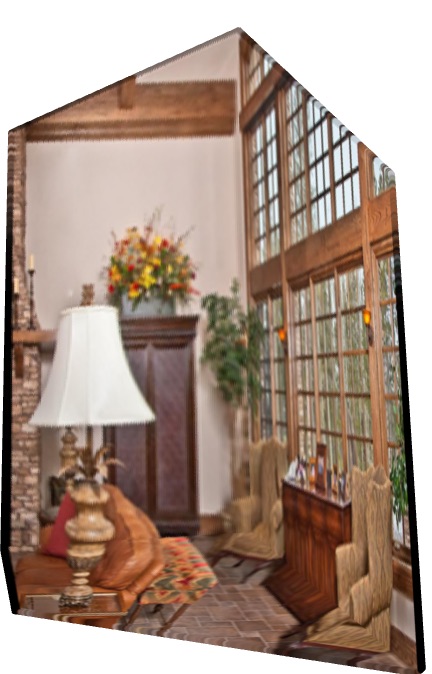}}&
\makecell{\includegraphics[width=0.24\linewidth,height=0.16\linewidth]{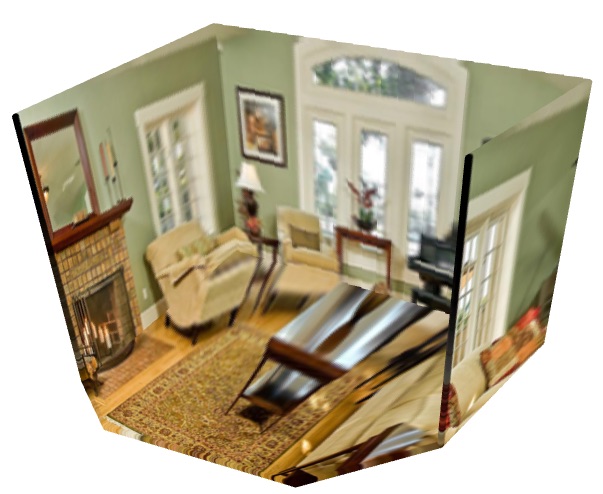}}&
\makecell{\includegraphics[width=0.24\linewidth,height=0.16\linewidth]{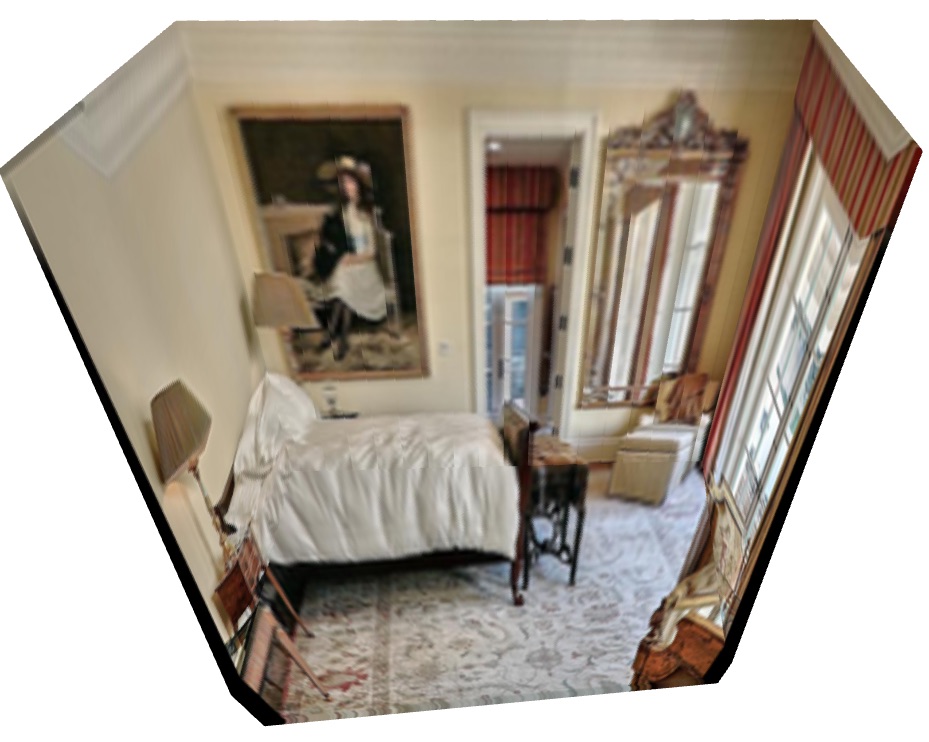}}&
\makecell{\includegraphics[width=0.24\linewidth,height=0.16\linewidth]{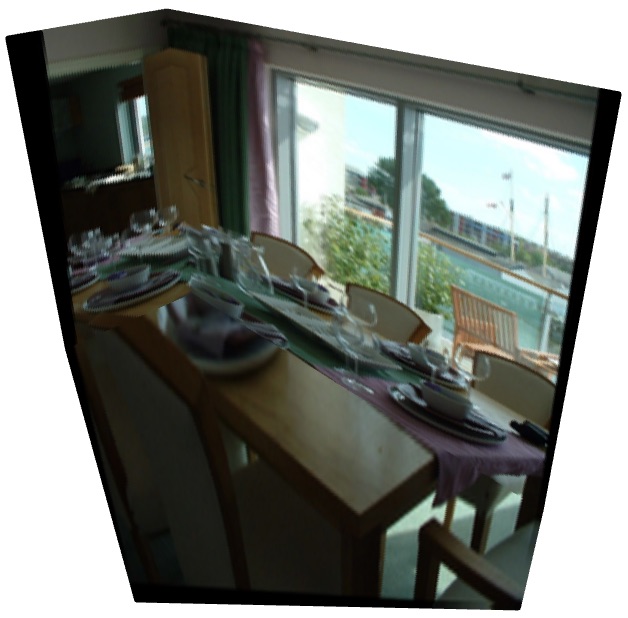}}
\end{tabular}
    
    \caption{Qualitative results on the LSUN validation set. The results are separately sampled from four groups that comprise the predictions with the best 0--25\%, 25--50\%, 50--75\% and 75--100\% corner errors (displayed from the first to the fourth column). The green lines are the ground-truth layout while the red lines are the estimated.}
    \label{fig:qual_lsun}
\end{figure*}

%% file: supp_qual_general.tex
\section{Qualitative Results for General Layout Topology}

\begin{figure*}[htb]
   \centering
   \setlength\tabcolsep{1pt}

\begin{tabular}{cccc}

(a) & (b) & (c) & (d) \\

\makecell{\includegraphics[width=0.24\linewidth,height=0.16\linewidth]{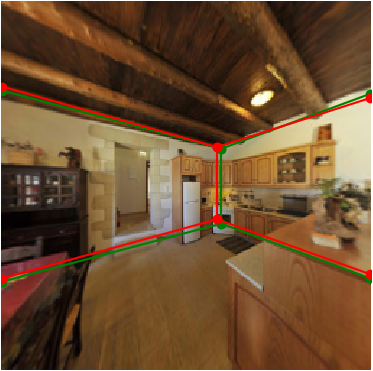}}&
\makecell{\includegraphics[width=0.24\linewidth,height=0.16\linewidth]{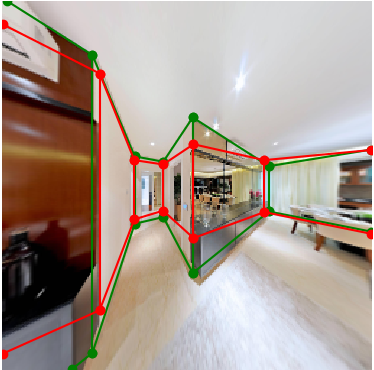}}&
\makecell{\includegraphics[width=0.24\linewidth,height=0.16\linewidth]{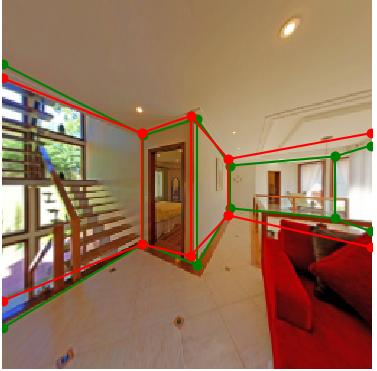}}&
\makecell{\includegraphics[width=0.24\linewidth,height=0.16\linewidth]{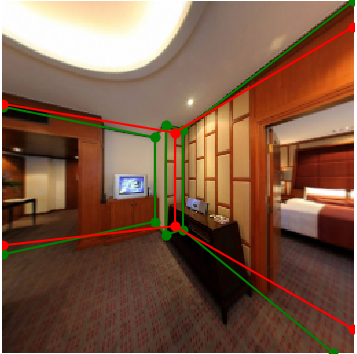}}
\\
\makecell{\includegraphics[width=0.24\linewidth,height=0.16\linewidth]{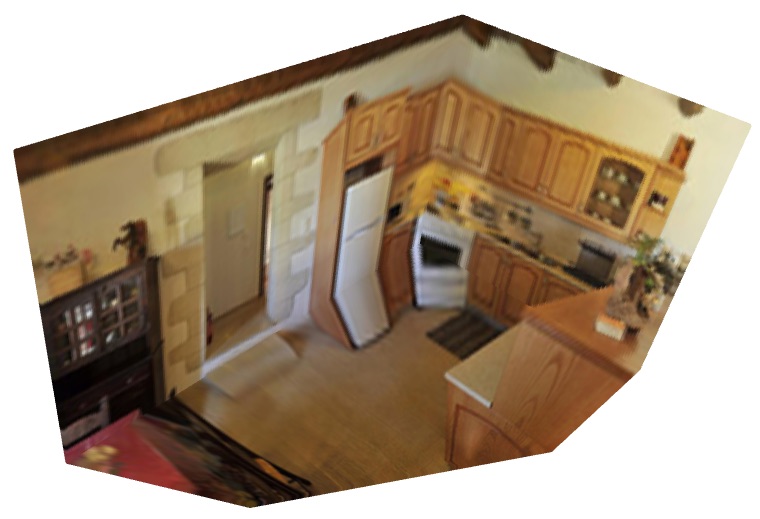}}&
\makecell{\includegraphics[width=0.24\linewidth,height=0.16\linewidth]{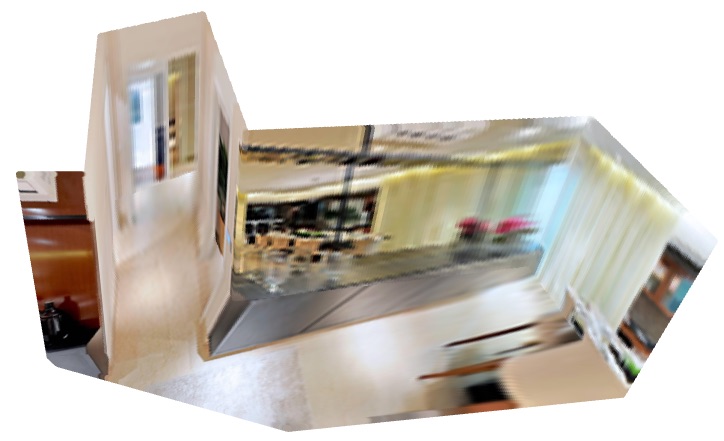}}&
\makecell{\includegraphics[width=0.24\linewidth,height=0.16\linewidth]{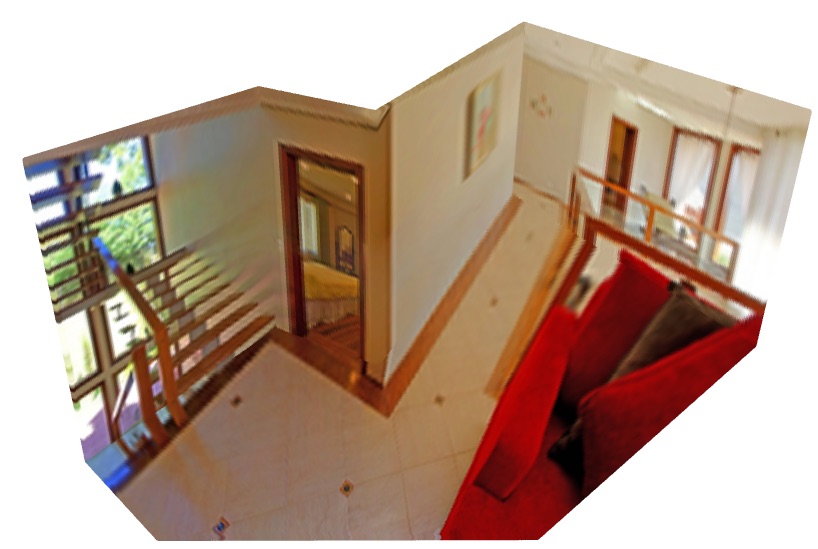}}&
\makecell{\includegraphics[width=0.24\linewidth,height=0.16\linewidth]{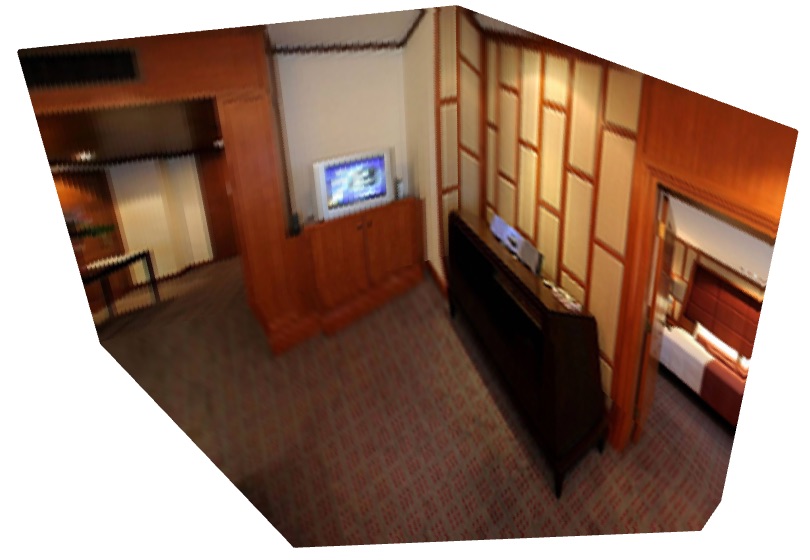}}
\\
\\
\makecell{\includegraphics[width=0.24\linewidth,height=0.16\linewidth]{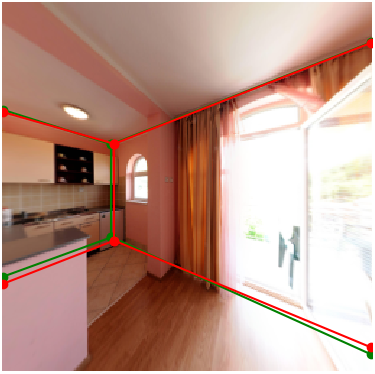}}&
\makecell{\includegraphics[width=0.24\linewidth,height=0.16\linewidth]{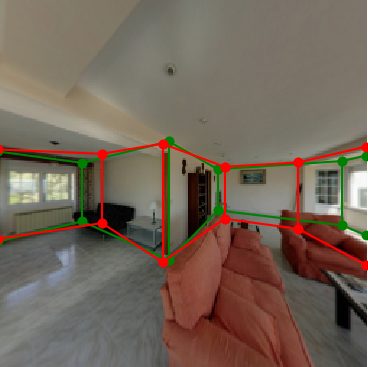}}&
\makecell{\includegraphics[width=0.24\linewidth,height=0.16\linewidth]{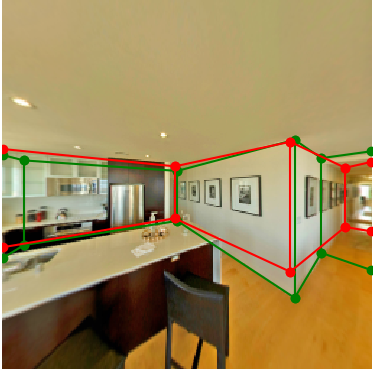}}&
\makecell{\includegraphics[width=0.24\linewidth,height=0.16\linewidth]{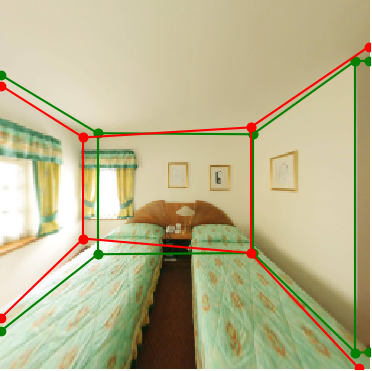}}
\\
\makecell{\includegraphics[width=0.24\linewidth,height=0.16\linewidth]{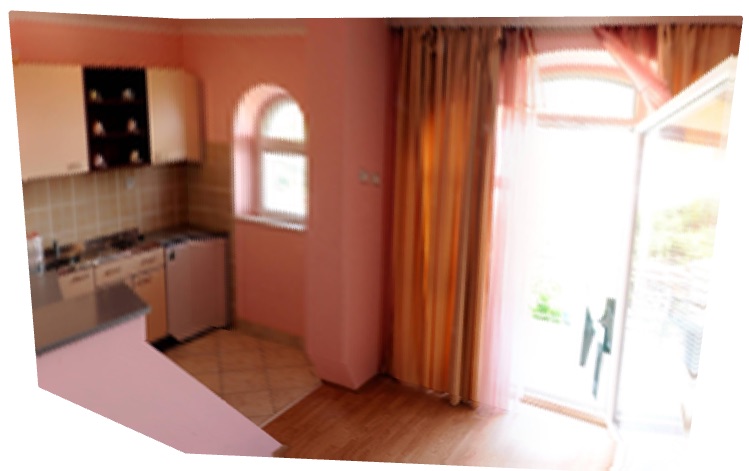}}&
\makecell{\includegraphics[width=0.24\linewidth,height=0.16\linewidth]{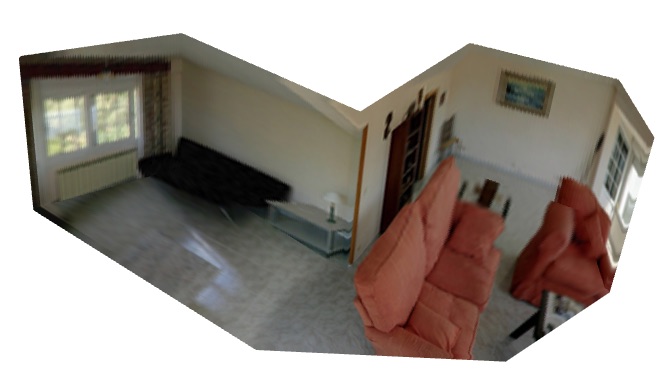}}&
\makecell{\includegraphics[width=0.24\linewidth,height=0.16\linewidth]{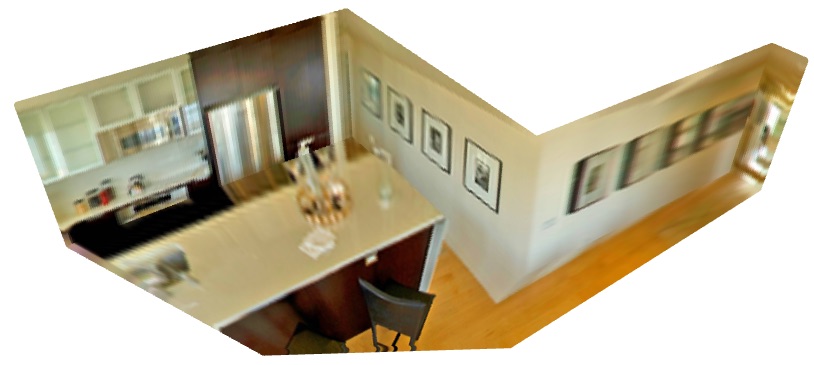}}&
\makecell{\includegraphics[width=0.24\linewidth,height=0.16\linewidth]{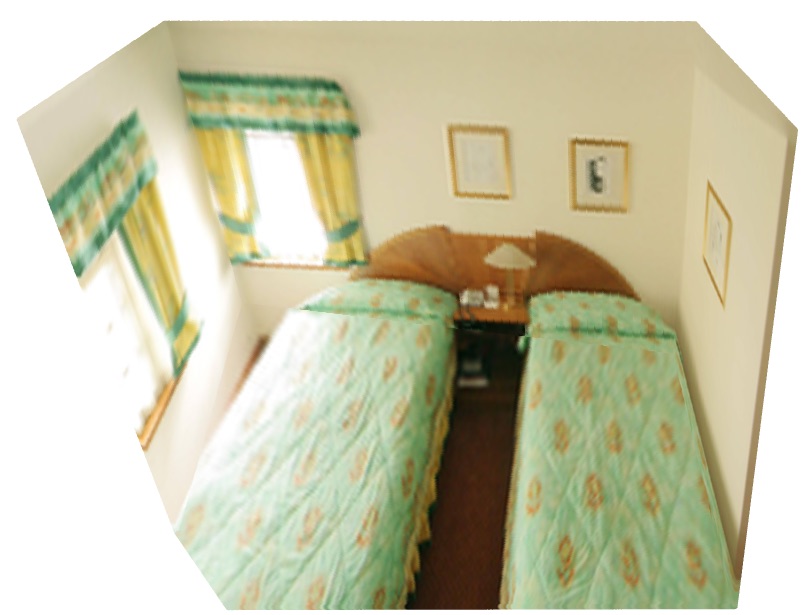}}
\\
\\
\makecell{\includegraphics[width=0.24\linewidth,height=0.16\linewidth]{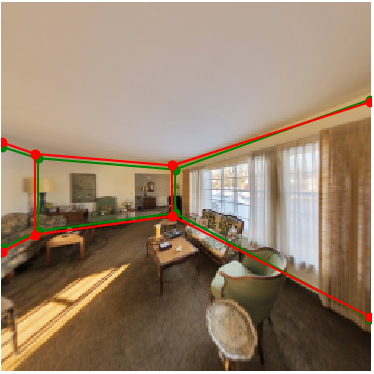}}&
\makecell{\includegraphics[width=0.24\linewidth,height=0.16\linewidth]{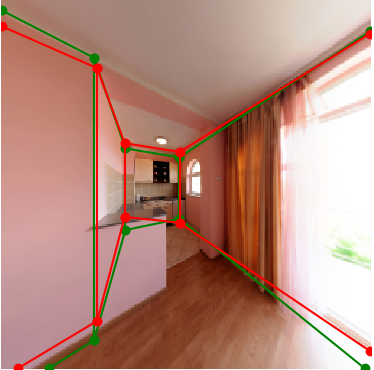}}&
\makecell{\includegraphics[width=0.24\linewidth,height=0.16\linewidth]{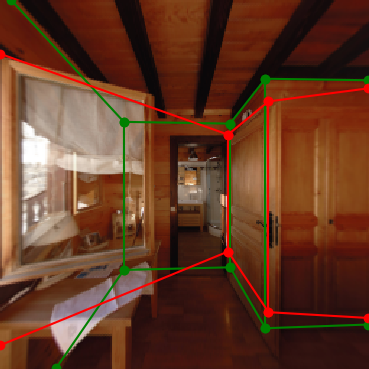}}&
\makecell{\includegraphics[width=0.24\linewidth,height=0.16\linewidth]{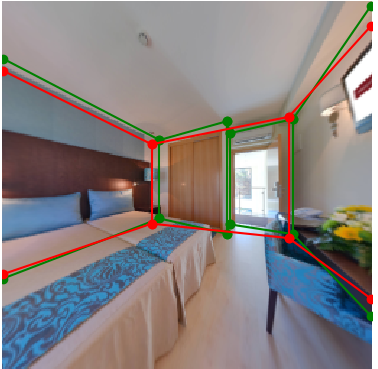}}
\\
\makecell{\includegraphics[width=0.24\linewidth,height=0.16\linewidth]{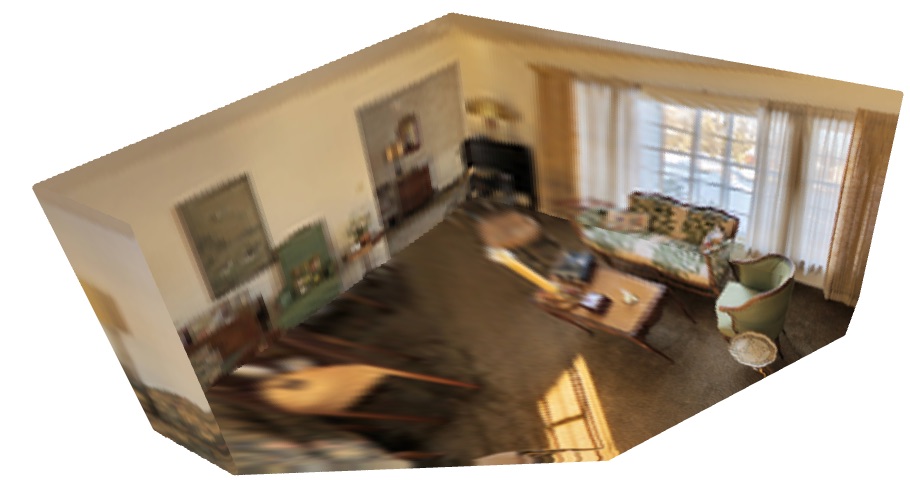}}&
\makecell{\includegraphics[width=0.24\linewidth,height=0.16\linewidth]{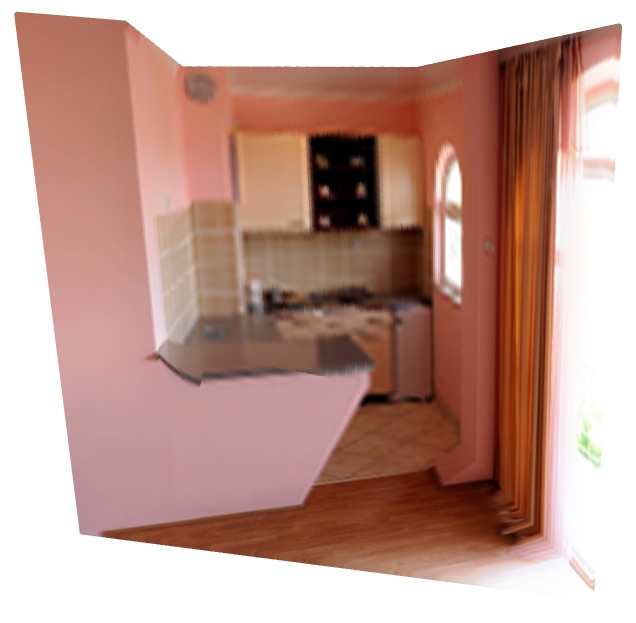}}&
\makecell{\includegraphics[width=0.24\linewidth,height=0.16\linewidth]{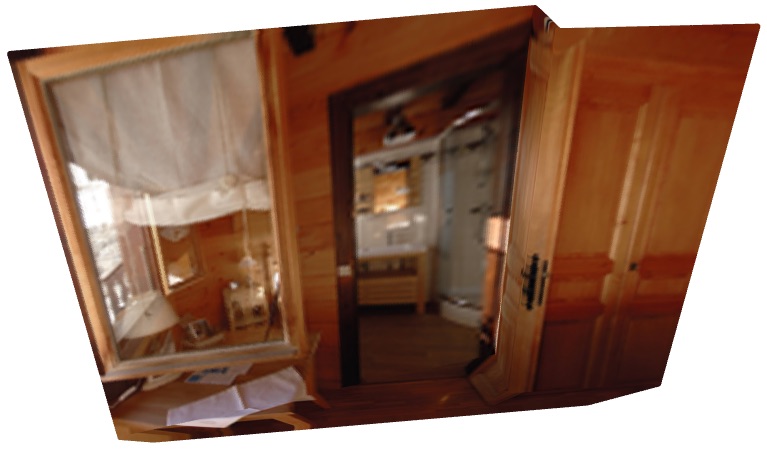}}&
\makecell{\includegraphics[width=0.24\linewidth,height=0.16\linewidth]{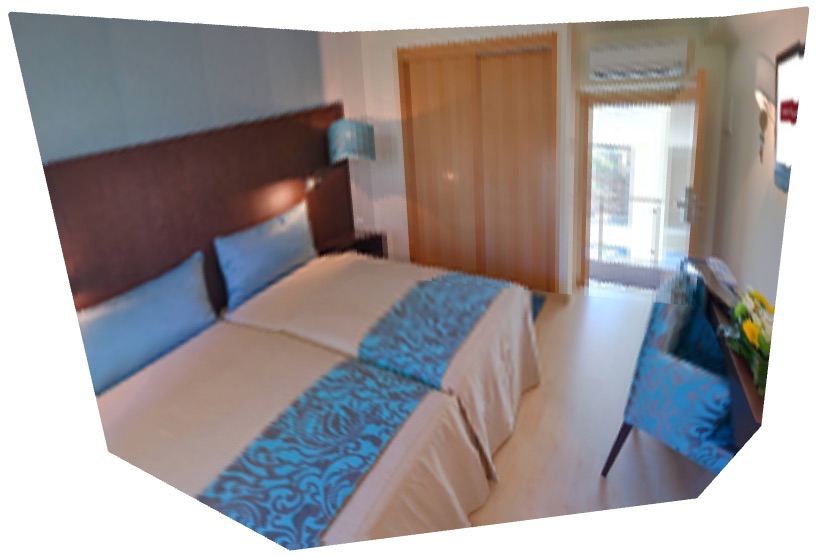}}
\end{tabular}
    
    \caption{Qualitative results for general layout topology. The results are separately sampled from four groups that comprise the predictions with the best 0--25\%, 25--50\%, 50--75\% and 75--100\% corner errors (displayed from the first to the fourth column). The green lines are the ground-truth layout while the red lines are the estimated.}
    \label{fig:qual_lsun}
\end{figure*}